\documentclass[10pt,twocolumn,letterpaper]{article}

 \usepackage{cvpr}              

\usepackage[dvipsnames]{xcolor}

\definecolor{cvprblue}{rgb}{0.21,0.49,0.74}
\usepackage[pagebackref,breaklinks,colorlinks,citecolor=cvprblue]{hyperref}
\usepackage[accsupp]{axessibility} 

\usepackage{graphicx}
\usepackage{amsmath}
\usepackage{amsfonts}
\usepackage{amsthm}
\usepackage{makecell}
\usepackage{diagbox}
\usepackage{xcolor}
\usepackage{booktabs}
\usepackage{algorithmic,float}
\usepackage[ruled,vlined]{algorithm2e}
\usepackage[switch]{lineno}

\usepackage{mathtools}
\usepackage{multirow}
\usepackage{subcaption}
\usepackage{stfloats}
\usepackage{footnote}

\def\paperID{17114} 
\def\confName{CVPR}
\def\confYear{2024}

\title{Strong Transferable Adversarial Attacks via Ensembled Asymptotically Normal Distribution Learning}

\author{Zhengwei Fang$^{1,2}$, Rui Wang$^{1,2,3}$\thanks{Corresponding author.}, Tao Huang$^{1,2}$, Liping Jing$^{1,2}$\\
$^1$School of Computer Science and Technology, Beijing Jiaotong University, Beijing, China\\
$^2$Beijing Key Lab of Traffic Data Analysis and Mining, Beijing, China\\
$^3$Collaborative Innovation Center of Railway Traffic Safety, Beijing, China\\
{\tt\small jankinfmail@gmail.com, \{rui.wang,thuang,lpjing\}@bjtu.edu.cn}\\
}

\begin{document}
\maketitle

\begin{abstract}
Strong adversarial examples are crucial for evaluating and enhancing the robustness of deep neural networks. However, the performance of popular attacks is usually sensitive, for instance, to minor image transformations, stemming from limited information — typically only one input example, a handful of white-box source models, and undefined defense strategies. Hence, the crafted adversarial examples are prone to overfit the source model, which hampers their transferability to unknown architectures. In this paper, we propose an approach named Multiple Asymptotically Normal Distribution Attacks (MultiANDA) which explicitly characterize adversarial perturbations from a learned distribution. Specifically, we approximate the posterior distribution over the perturbations by taking advantage of the asymptotic normality property of stochastic gradient ascent (SGA), then employ the deep ensemble strategy as an effective proxy for Bayesian marginalization in this process, aiming to estimate a mixture of Gaussians that facilitates a more thorough exploration of the potential optimization space. The approximated posterior essentially describes the stationary distribution of SGA iterations, which captures the geometric information around the local optimum. Thus, MultiANDA allows drawing an unlimited number of adversarial perturbations for each input and reliably maintains the transferability. Our proposed method outperforms ten state-of-the-art black-box attacks on deep learning models with or without defenses through extensive experiments on seven normally trained and seven defense models. 
\end{abstract}

\section{Introduction}

\label{sec:intro}

\begin{figure}[ht]
    \centering
    \includegraphics[width=0.98\linewidth]{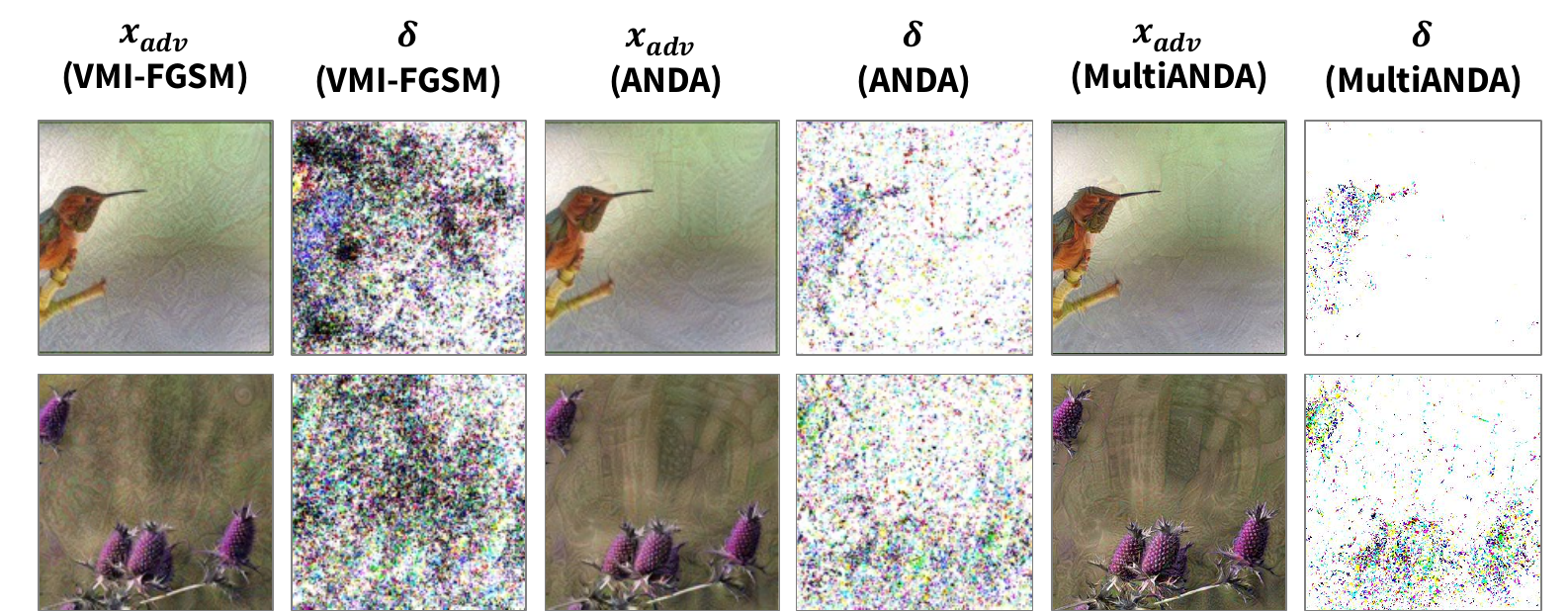}
    \caption{Perturbation visualization of the adversarial examples generated by ANDA/MultiANDA that fool all selected normally trained (first row) and defense (second row) models.}
    \label{fig:vis_main}
\end{figure}

Albeit the excellent performance of deep neural networks (DNNs) across a broad spectrum of tasks in machine vision and natural language processing, a rich stream of work \cite{SZE14,Goodfellow15,Dong18,Zhang22,Zhao22} shows the non-negligible vulnerability of DNNs to adversarial attacks, which craft imperceptible perturbations to synthesize malicious inputs for triggering unexpected model predictions. Such severe degradation of model robustness and stability greatly hinders the application of DNNs in security- or safety-critical domains \cite{AMO16,ALV18,FAR18}. Fortunately, studying how to create potent adversarial examples aids in evaluating DNN-based systems and enhancing their robustness by devising defensive strategies, such as adversarial training \cite{Goodfellow15,Madry18}.

This paper focuses on the black-box and transfer-based threat model. The corresponding adversarial attack algorithms intend to synthesize the adversarial examples that can successfully fool the unknown DNNs without accessing any information on model architectures and outputs. Researchers implement such attacks by maximizing the non-concave loss function of a substitute model trained for the same task with, for example, one step \cite{Goodfellow15}, or iterative steps \cite{Kurakin16} of gradient ascent optimization. However, prior work has shown the generated adversarial examples trend to overfit the substitute model, which hardly attacks other models \cite{Kurakin17}. Meanwhile, they are often sensitive to minor transformations \cite{Luo15,Lu17,Athalye18}.

Hence, researchers have been exploring various data augmentation strategies \cite{Athalye18,Xie19,Dong19,Lin20} and improved SGA optimization methods \cite{Dong18,Wang21a,Huang2022transferable,Lin20} to enhance the performance of transferable attacks. Nonetheless, these efforts exhibit limited generalization capabilities on unknown DNN architectures, particularly those trained with defense strategies.
This suggests that a deterministic optimization procedure, initiated from a single image, is insufficient to thoroughly explore the high-dimensional perturbation space.
Furthermore, the projected gradient descent (PGD) method proposed by Madry \etal \cite{Madry18} initializes the attack algorithm with random restarts around the legitimate image. However, these random samples represent homogeneous deviations from the original input, lacking diversity \cite{Dong20}, which limits the transferability of adversarial examples.

To resolve these issues, in this paper, we propose \textbf{Multi}ple \textbf{A}symptotically \textbf{N}ormal \textbf{D}istribution \textbf{A}ttacks (\textbf{MultiANDA}), a novel method that explicitly characterizes perturbations inferred from a learned distribution.
Firstly, we formulate a single ANDA by taking advantage of the nice property of stochastic gradient ascent (SGA), the asymptotic normality \cite{Chen16,Mandt17,Izmailov18,Maddox19}, to learn the true optimal posterior distribution of adversarial perturbations. 
Specifically, ANDA calculates and stores the first two moments of iterations captured along the optimizing trajectory to approximate the distribution of the adversarial perturbation as a Gaussian one.
Moreover, we apply the deep ensemble strategy \cite{LAK17} on ANDA to devise a mixture of Gaussians, approximating the Bayesian posterior. As each Gaussians positioned at a different basin of attraction, the proposed MultiANDA leads to a further improved generalization performance of the attacks with more diverse adversarial examples. As shown in Figure~\ref{fig:vis_main}, the perturbations crafted by our proposed method focus more on the semantic and decisive areas of objects than the baseline approach.
In particular, the contributions of this work are the following:

\textbullet\  We propose MultiANDA, an approximate Bayesian inference method to realize transfer-based adversarial attacks. We learn the perturbation distribution instead of finding a specific adversary for each input example. Thus, a good approximation of the true perturbation posterior boosts the generalization (i.e., transferability) of the generated adversarial examples.

\textbullet\  We devise a stochastic adversarial attacking process instead of the original deterministic one initialized from one input by random image translations. In this way, we profit the asymptotic normality of SGA for a better perturbation distribution learning. Adopting the deep ensemble strategy, an effective mechanism for approximating Bayesian marginalization \cite{WIL20}, MultiANDA accurately and efficiently estimate the posterior distribution over the perturbations by using the geometry information along the attacking trajectory.

\textbullet\  The learned distributions of perturbation allow drawing an unlimited number of adversarial examples for one input. We experimentally validate that the sampled adversaries have comparably high attack success rates with the singly generated ones, even on the defense models. This result demonstrates the great potential of our method for designing defense strategies grounded in adversarial training.

\textbullet\  Extensive experiment results show that the proposed method outperforms ten state-of-the-art black-box attacks on deep learning models with or without advanced defenses. Significantly, MultiANDA is more advantageous when attacking the advanced defense models. Thus, ANDA and MultiANDA are expected to benchmark new defense methods in the future. Our code is available at \url{https://github.com/CLIAgroup/ANDA}.  

\section{Related Work and Preliminaries}
\label{sec:relat}
Our study mainly focuses on the black-box and transfer-based threat model, as they may bring more security and safety risks due to their practical applicability \cite{Papernot17,Dong18}.

\subsection{Black-Box Adversarial Attacks}
\label{sec:bbattack}
Essentially, the adversarial attack algorithms intend to maximize the loss function $\mathcal{L}(\cdot)$, e.g., the cross entropy loss for classification tasks, to search for the adversaries $x_{adv}$ \cite{SZE14,Goodfellow15}, with the constraints of $l_\infty$ norm bound $\varepsilon$, as shown in (\ref{eq:obj_fgsm}):
\begin{equation}\label{eq:obj_fgsm}
    \max_{x_{adv}} \mathcal{L}(x_{adv},y) \quad  \mathrm{s.t. ~} ||x_{adv}-x||_\infty\leq\varepsilon,
\end{equation}
where $x$, $x_{adv}$, $y$ are the legitimate input, the adversarial example and the true label, respectively.

Among prior work, the fast gradient sign method (FGSM) \cite{Goodfellow15}  and its iterative version, the basic iterative method (BIM) \cite{Kurakin16} resort to linearizing the non-concave loss function and realize the attack with one-step update or iterative updates, as shown in (\ref{eq:algo_bim}):
\begin{equation}\label{eq:algo_bim}
    x_{adv}^{(t+1)}=\Phi(x_{adv}^{(t)} + \alpha \cdot \operatorname{sign}(\underbrace{\nabla_x \mathcal{L}(x_{adv}^{(t)},y)}_{\delta^{(t)}})),
\end{equation}
where $x_{adv}^{(t+1)}$, $\delta^{(t)}$ are the adversarial example and perturbation at the $t$-th step,  $x_{adv}^{(0)}$ is the legitimate input, and $\Phi(\cdot)$ is the projection function on the $\varepsilon$-ball.

These attacks are more effective under white-box scenarios, whereas they tend to overfit the model parameters and hardly transfer to unknown architectures \cite{Kurakin17}.
Thus, many works have been proposed to improve the transferability of adversarial examples, which can be divided into three categories. The first group focuses on the iterative optimizing process by considering momentum~\cite{Dong18}, Nesterov accelerated gradient~\cite{Lin20}, and variance reducing~\cite{Wang21a} to search for the optimal solutions.
The second group attempts to attack the mid-layers by disrupting the critical object-aware features~\cite{Zhou18,Naseer2018task,Ganeshan19,Huang19,Wang21b}

Moreover, the third stream of methods adopt the commonly used generalization techniques by introducing diversity. These techniques encompass data augmentation \cite{Krizhevsky12,He16}, such as resize and padding \cite{Xie19}, translation \cite{Dong19}; and other integrated augmentation strategies, including combining multiple linear transformations \cite{Athalye18}, modifying the pixel values with multiple scales \cite{Lin20}, integrating more information from other images~\cite{Cohen21}, and using the sum of the gradients from intensity augmented images~\cite{Huang2022transferable}.

Similar to our study, existing attack approaches also try to determine the explicit adversarial distributions. PGD \cite{Madry18} resolves the inner maximization problem by the random restart strategy. However, the uniformly drawn samples tend to lie densely around the natural image \cite{Dong20}. Li \etal \cite{Li19} propose $\mathcal{N}$attack to define and learn an adversarial distribution. However, their work is under a query-based threat model, i.e., it has to access the output scores for each input from the black-box defense model, which weakens its scalability in real-world applications.
Additionally, Dong et al. \cite{Dong20} propose to formulate the adversarial perturbations via a learned distribution for adversarial distributional training (ADT). Similar to $\mathcal{N}$attack, ADT crafts the adversarial examples centered with the natural input $x$, which may fail in generating adversarial examples once the input is a little bit far away from the boundary. Conversely, ANDA generates adversaries around the optimal $x^*_{adv}$. In this case, the adversaries are more prone to successful attacks. From the facets of implementation, ADT focuses on learning a distribution from a large amount of training data for adversarial training. Our proposed method tries to collect the trajectory information during the optimization process for each natural input. Thus, these crucial differences make two approaches suitable for distinct tasks: inner maximization problem for adversarial training (ADT) and transfer-based black-box attack (ANDA and MultiANDA).

\subsection{Asymptotic Normality of SGD}

Maddox et al. \cite{Maddox19} point out that the trajectory of stochastic gradient descent (SGD) can be regarded as the Bayesian posterior distribution over the random variables to be optimized, e.g., the model parameters in the case of the DNN training. Following this theory, they propose a Gaussian posterior approximation method (SWAG) based on the first two moments of SGD iterations for model calibration or uncertainty quantification. In the prior work, Mandt et al. \cite{Mandt17} provide a more detailed analysis of the stationary distribution of SGD iterations with a constant learning rate, also termed the asymptotic normality of SGD \cite{Van98}. Such analysis can be traced back to the work by Ruppert \cite{Ruppert88}, Polyak and Juditsky \cite{Polyak92} known as Polyak-Ruppert averaging, and still an active research subject for SGD \cite{Chen16}. Furthermore, inspired by deep ensembles \cite{LAK17}, Wilson and Izmailov \cite{WIL20} carried out an extensive empirical study, demonstrating that applying SWAG only with multiple randomly initialized models can significantly improve the model performance. They emphasize that this benefits from the multimodal effects of the mixture Gaussian distributions and, in turn, boosts the contribution of each additional sample in the optimization procedure.
In this paper, we adopt these theories to innovatively model the posterior distribution of perturbations to search for potent adversarial examples. 

\section{Proposed Method}
\label{sec:metho}

Drawing on the comprehensive research previously discussed, we hypothesize that utilizing the asymptotic normality of SGD (SGA in our problem formulation) to marginalize an approximate Bayesian posterior could enhance generalization (transferability in our context) beyond what is achievable with a fixed  set of parameter (one specific adversarial perturbation). However, optimizing the objective (\ref{eq:obj_fgsm}) using attack algorithms such as BIM (\ref{eq:algo_bim}) is a deterministic gradient ascent process. In this process, no gradient noise is introduced to form a stationary distribution of SGD. Thus, to bridge this gap, we exploit random data augmentation in the standard optimization process, devising a novel attack method for accurately and efficiently estimating the posterior distribution over perturbations.

In Section \ref{sec:condi}, we first empirically analyze the asymptotically normal distribution property in the stochastic adversarial attack process, then proceed to formalize the optimization objective for the task of adversarial example generation. Finally, inspired by the advantageous properties of SGD and the effective multimodal inference of the ensemble strategy, we introduce the new attack methods ANDA and MultiANDA in Sections \ref{sec:anda} and \ref{sec:multianda}, respectively.

\subsection{Asymptotic Normality of Stochastic BIM}
\label{sec:condi}

Under specific conditions—decaying learning rates, smooth gradients, and a full rank stationary distribution—it has been theoretically established that SGD asymptotically converges to a Gaussian distribution centered at the local optimum \cite{Asmussen07,Chen16,Maddox19}. Expanding on this, Mandt et al. \cite{Mandt17} demonstrated that under certain assumptions, SGD with a constant and sufficiently small learning rate in its final search phase, i.e., when updates are proximal to the local optimum, converges to a stationary distribution. This convergence implies that iterations during this phase can approximate the posterior distribution of the optimized variables. Crucially, the gradient noise, introduced by the randomness of the sampled training batches, is instrumental in forming this distribution, provided the learning rate is adequately small \cite{Maddox19}. These theoretical underpinnings motivate our approach in the algorithm design.

\begin{figure}[ht]
    \centering
    \includegraphics[width=1\linewidth]{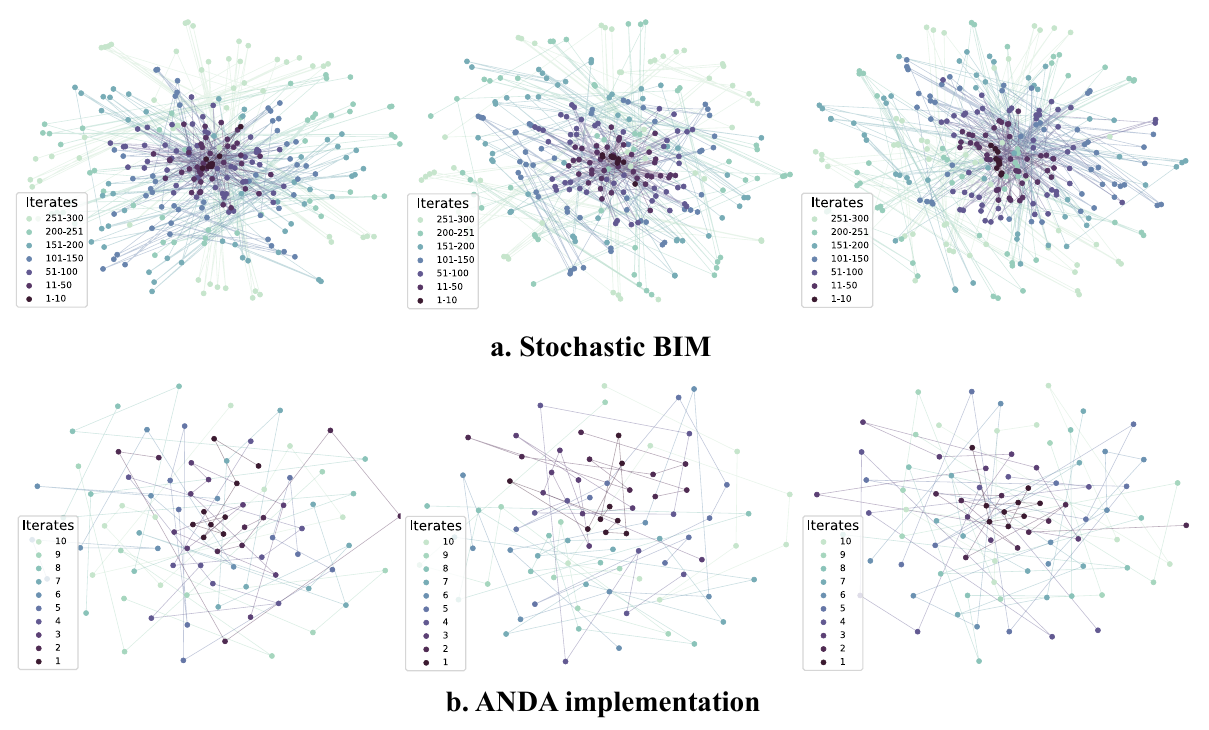}
    \vspace{-0.5cm}
    \caption{Asymptotically normal distribution examples with iterative trajectories of adversarial perturbations generated by stochastic BIM (a) and the proposed approach (b). }
    \label{fig:Gaussians}
\end{figure}



As demonstrated in (\ref{eq:algo_bim}), the optimization process, beginning from a specific input, is a deterministic gradient ascent, differing from a standard SGA approach. Therefore, we introduce stochastic gradient noises to emulate Gaussian limiting distributions by integrating a random data augmentation technique $\texttt{AUG}$, like image translation used in our experiments, at each iterative step:
\begin{equation}\label{eq:algo_stochastic_bim}
    x_{adv}^{(t+1)}=\Phi(x_{adv}^{(t)} + \alpha \cdot \operatorname{sign}(\underbrace{\nabla_x \mathcal{L}(\texttt{AUG}(x_{adv}^{(t)}),y)}_{\delta^{(t)}})).
\end{equation}

Notably, a small constant learning rate schedule ($\alpha=\epsilon/T$) is commonly adopted, where $\epsilon \in \mathbb{R} $ is a sufficiently small number used to restrict the $L_{\infty}$-norm for creating invisible adversarial perturbations, and $T$ is the number of iterative steps. Hence, we speculate the devised stochastic BIM shown in (\ref{eq:algo_stochastic_bim}) can be used as an approximate Bayesian inference algorithm. In particular, Mandt et al. \cite{Mandt17} assume that the gradients along the optimizing trajectory follow the Gaussian distribution stemming from the central limit theorem. We visualized the adversarial perturbations $\delta^{(t)}$—the loss gradients in our task—using the t-SNE technique \cite{Maa08} to examine their distribution. 300 iterative steps for 3 randomly-drawn ImageNet \cite{Rus15} images shown in Fig.~\ref{fig:Gaussians}~\textbf{a} illustrates the Gaussian-shaped distributions which underpins our conjecture
of the asymptotically normality of stochastic BIM algorithm. We notice the obvious increased gradient noise along the optimizing trajectory, which may due to the use of the $\operatorname{sign}()$ function and projection operations. Nevertheless, this does not affect the convergence of perturbations. Importantly, the first 10 steps converging with tight covariance justifies another speculation that the small value of $T$ (often =10 for non-targeted attacks) and the non-trivial first step in our task (e.g., referring to strong attack performance of FGSM) indicate that the early iterates of are fairly close to the local optimum (i.e., the final search phase discussed above).

Inspired by the results of this empirical study, we propose employing $n$ image transformations $\texttt{AUG}_i(\cdot)$ to form a minibatch for an adversarial attack, thereby further enhancing stochasticity. Then, the optimization objective is formulated as
\begin{equation}\label{eq:ori_obj}
    \max_{x_{adv}} \sum_{i=1}^n\mathcal{L}(\texttt{AUG}_i(x_{adv}),y) \quad  \mathrm{s.t. ~} ||x_{adv}-x||_\infty\leq\varepsilon.
\end{equation}
Furthermore, optimizing (\ref{eq:ori_obj}) to craft just one optimal adversarial example does not sufficiently explore the high-dimensional potential perturbation space. To further improve the transferability of adversarial examples to unknown architectures, we propose to explicitly model the adversarial perturbation ($\delta$) via a distribution ($\Pi_{\delta}$). 
Thus, our optimization objective is formalized as
\begin{equation} \label{eq:our_obj}
    \begin{gathered}
        \max_{\Pi_{\delta}} \mathbb{E}_{\delta \sim \Pi_{\delta}} \sum_{i=1}^n\mathcal{L}(\texttt{AUG}_i(x + \delta),y),
        \mathrm{s.t. ~} ||\delta||_\infty\leq\varepsilon.
    \end{gathered}
\end{equation}
Objective (\ref{eq:our_obj}) aims to maximize the expected loss over the adversarial perturbation distribution. It is noteworthy that (\ref{eq:our_obj}) is a generalized version of (\ref{eq:ori_obj}) and is expected to characterize sufficient information about perturbations for transferable adversarial example generation.

\begin{algorithm}[tb]\small
    \caption{Asymptotically Normal Distribution Attack (ANDA)}
    \setlength{\baselineskip}{9pt}
    \label{alg:example}
    \textbf{Input:} $x$: clean image $x \in \mathbb{R}^{d}$; $y$: ground-truth label; $f$: pre-trained source model;
    \vspace{0.1cm}

    \textbf{Parameters:} $T$: \# iterations; $\epsilon$: perturbation magnitude; $n$: \# batch samples for augmentation; $M$: \# sampling examples;
\vspace{0.1cm}

    \textbf{Output:} $x_{adv}$: Adversarial examples; 

    \begin{algorithmic}[1]
        \STATE \textbf{Initialize } $\alpha \leftarrow \epsilon / T $,  $\bar{\delta}^{(0)} \leftarrow \mathbf{0}^d $,  $\mathbf{D} \leftarrow \emptyset $, $x_0 \leftarrow x$
        \FOR{$t=0$ {\bfseries to} $T-1$}
        \STATE Calculate the mean of gradients and store the deviation of gradients throughout previous $t$ iterations:
        \begin{equation*}
            \begin{aligned}
                 & \delta_i^{(t)} = \nabla_{x^{(t)}}{J(f(\texttt{AUG}_i(x^{(t)})),y)}, i = 1, \dots, n \\
            \end{aligned}
        \end{equation*}

        \begin{equation*}
            \begin{aligned}
                 & \bar{\delta}^{(t+1)} = \frac{(t \times n )\bar{\delta}^{(t)} +  \sum_{i=1}^n\delta_{i}^{(t)}}{(t+1)\times n} \\
            \end{aligned}
        \end{equation*}

        \begin{equation*}
            \begin{aligned}
                 & \texttt{APPEND}\_\texttt{COLS}( \mathbf{D}, \{ \delta_i^{(t)} - \bar{\delta}^{(t+1)} \} ), i =1, \dots, n \\
            \end{aligned}
        \end{equation*}

        \STATE Update the adversarial example along the average direction:
        \begin{equation*}
            \begin{aligned}
                 & x^{(t+1)} = \texttt{Clip}_{x, \epsilon} \{ x^{(t)} + \alpha \cdot \texttt{Sign}(\bar{\delta}^{(t+1)})\}
            \end{aligned}
        \end{equation*}
        \ENDFOR

        \STATE  Output option (a): Craft one $x_{adv}$\\

            \begin{equation*}
                \begin{aligned}
                     & x_{adv} = x^{(T)} \\
                \end{aligned}
            \end{equation*}
            Output option (b): Craft M $x_{adv}$ by sampling from the learned perturbation distribution $\mathcal{N}(\bar{\delta},\sigma)$\\
            
            \begin{equation*}
                \begin{aligned}
                     & C = \texttt{NUMS}\_\texttt{COLS}(\mathbf{D}) = n \times T                            \\
                     & \bar{\delta} = \bar{\delta}^{(T)}, \quad \sigma = \frac{\mathbf{D}\mathbf{D}^T}{C-1} \\
                \end{aligned}
            \end{equation*}
             Generate adversarial examples \\ \vspace{-0.3cm}
            \begin{equation*}
                \begin{aligned}
                     & \{\delta_{m}\}_{m=1}^{M}  \sim \mathcal{N}\left(\bar{\delta}, \sigma \right),                                               \\
                     & \{x_{adv}^{m}\}_{m=1}^{M} = \{\texttt{Clip}_{x, \epsilon} \{ x^{(T-1)} + \alpha \cdot \texttt{Sign}(\delta_m)\}\}_{m=1}^{M}
                \end{aligned}
            \end{equation*}

    \end{algorithmic}
\end{algorithm}

\begin{table*}[!t]
    \footnotesize
    \centering
    \vspace{0.5cm}
    \setlength\tabcolsep{4pt}
    \begin{tabular}{c|cccccc|cccccc}
        \toprule
        \multirow{2}{*}{Attack} & \multicolumn{6}{c|}{Inc-v3 $\Longrightarrow$} & \multicolumn{6}{c}{ResNet-50 $\Longrightarrow$} \\
                   & Inc-v3            & ResNet-50        & ResNet-152    & IncRes-v2        & VGG-19                 & Avg.              & Inc-v3            & ResNet-50         & ResNet-152        & IncRes-v2        & VGG-19            & Avg.          \\ \hline
        BIM        & \textbf{100.0*}   & 20.3             & 15.7                & 15.6             & 34.3             & 21.5              & 33.2              & 99.7*             & 63.9              & 20.9             & 43.4              & 40.4          \\
        TIM        & 64.3*             & 35.9             & 30.6                & 25.4             & 70.4             & 45.3              & 47.3              & 77.0*             & 47.6              & 30.8             & 68.8              & 54.3          \\
        SIM        & \textbf{100.0*}   & 38.2             & 31.1                & 35.9             & 42.2             & 36.9              & 48.6              & \textbf{100.0*}   & 83.7              & 35.3             & 50.9              & 54.6           \\
        DIM        & \textbf{100.0*}   & 31.7             & 25.5                & 31.4             & 45.5             & 33.5              & 61.9              & \underline{99.9*} & 83.6              & 47.5             & 61.1              & 63.5           \\
        FIA        & 98.3*             & \underline{78.4} & \underline{75.3}    & 81.2             & \textbf{83.5}    & \textbf{79.6}     & 86.2              & 99.6*             & 94.5              & 80.4             & 88.3              & 87.4           \\
        TAIG       & \underline{99.7*} & 53.3             & 45.9                & 56.7             & 54.2             & 52.5              & 62.4              & \textbf{100.0*}   & 86.1              & 51.9             & 58.7              & 64.8           \\
        NI-FGSM    & \textbf{100.0*}   & 40.0             & 35.2                & 39.9             & 56.9             & 43.0              & 59.6              & 99.7*             & 85.4              & 48.1             & 67.0              & 65.0           \\
        MI-FGSM    & \textbf{100.0*}   & 40.2             & 35.1                & 40.3             & 57.1             & 43.2              & 58.5              & 99.7*             & 85.9              & 48.6             & 67.4              & 65.1           \\
        VMI-FGSM   & \textbf{100.0*}   & 63.0             & 59.3                & 68.6             & 70.3             & 65.3              & 75.3              & \underline{99.9*} & 93.4              & 68.3             & 76.4              & 78.4           \\
        VNI-FGSM   & \textbf{100.0*}   & 62.4             & 58.7                & 67.7             & 69.7             & 64.6              & 75.4              & 99.8*             & 92.9              & 67.9             & 75.6              & 78.0           \\\cline{1-13}
        ANDA       & \textbf{100.0*}   & 76.1             & 72.8                & \underline{82.3} & 77.0             & \underline{77.1}  & \underline{95.6}  & \textbf{100.0*}   & \underline{98.9}  & \underline{94.0} & \underline{89.5}  & \underline{94.5} \\
        MultiANDA  & \textbf{100.0*}   & \textbf{79.2}    & \textbf{76.0}       & \textbf{84.5}    & \underline{78.8} & \textbf{79.6}  & \textbf{96.5}     & \textbf{100.0*}   & \textbf{99.2}     & \textbf{95.0}    & \textbf{90.1}     & \textbf{95.2}    \\

        \hline
        \multirow{2}{*}{Attack} & \multicolumn{6}{c|}{IncRes-v2 $\Longrightarrow$}    & \multicolumn{6}{c}{VGG-19 $\Longrightarrow$} \\                                                                                                                          
                    & Inc-v3             & ResNet-50         & ResNet-152        & IncRes-v2         & VGG-19            & Avg.              & Inc-v3             & ResNet-50          & ResNet-152          & IncRes-v2         & VGG-19            & Avg.             \\ \hline
        BIM         & 36.3               & 25.6              & 20.6              & 99.3*             & 37.8              & 30.1              & 23.5               & 18.7               & 13.7                & 9.9               & \underline{99.9*} & 16.5              \\
        TIM         & 43.4               & 36.5              & 32.0              & 36.0*             & 66.2              & 42.8              & 41.5               & 34.8               & 30.2                & 23.6              & \textbf{100.0*}   & 46.0          \\
        SIM         & 58.9               & 47.8              & 41.4              & \underline{99.6*} & 49.8              & 49.5              & 37.7               & 34.4               & 25.2                & 23.6              & \textbf{100.0*}   & 30.2              \\
        DIM         & 54.6               & 41.4              & 36.6              & 98.2*             & 50.0              & 45.7              & 31.7               & 26.4               & 19.2                & 16.4              & \underline{99.9*} & 23.4              \\
        FIA         & 82.2               & 75.3              & 72.4              & 89.2*             & 80.7              & 77.7              & 57.4               & 50.7               & 40.9                & 42.7              & \textbf{100.0*}   & 47.9              \\
        TAIG        & 73.9               & 63.4              & 58.4              & 95.0*             & 57.4              & 63.3              & 48.8               & 43.6               & 34.7                & 33.9              & \textbf{100.0*}   & 40.3              \\
        NI-FGSM     & 61.9               & 49.5              & 44.7              & 99.2*             & 64.7              & 55.2              & 43.6               & 39.5               & 29.2                & 30.4              & \underline{99.9*} & 35.7              \\
        MI-FGSM     & 60.3               & 49.3              & 43.0              & 98.8*             & 64.6              & 54.3              & 44.7               & 39.4               & 28.9                & 30.8              & \underline{99.9*} & 36.0              \\
        VMI-FGSM    & 81.1               & 69.6              & 66.4              & 99.3*             & 73.5              & 72.7              & 62.7               & 56.7               & 46.5                & 48.6              & \textbf{100.0*}   & 53.6              \\
        VNI-FGSM    & 80.9               & 70.0              & 65.8              & 99.4*             & 73.8              & 72.6              & 63.2               & 56.7               & 46.6                & 48.8              & \textbf{100.0*}   & 53.8              \\\cline{1-13}
        ANDA        & \underline{93.0}   & \underline{86.4}  & \underline{83.7}  & \textbf{99.8*}    & \underline{82.8}  & \underline{86.5}  & \underline{74.4}   & \underline{64.1}   & \underline{56.4}    & \underline{61.5}  & \textbf{100.0*}   & \underline{64.1}  \\
        MultiANDA   & \textbf{93.9}      & \textbf{87.1}     & \textbf{85.6}     & \textbf{99.8*}    & \textbf{84.3}     & \textbf{87.7}     & \textbf{75.4}      & \textbf{66.1}      & \textbf{58.6}       & \textbf{63.5}     & \textbf{100.0*}   & \textbf{65.9}     \\

        \bottomrule
    \end{tabular}
    \caption{\label{tab:normaltrained}
       The success rates (\%) of the proposed and baseline attacks on five normally trained models. Sign * indicates the results on white-box source models. The best/second results are shown in bold/underlined. Due to the space limit, performance on five target models are presented here. See full results on seven targets in Appendix.
   }
\end{table*}

\subsection{Asymptotically Normal Distribution Attack}
\label{sec:anda}

To better solve the optimization problem formulated in (\ref{eq:our_obj}), we take advantage of the asymptotically normality of SGA and iteratively search for the adversarial examples similarly to (\ref{eq:algo_bim}).
We omit the projection function $\Phi(\cdot)$ and $\operatorname{sign}(\cdot)$ because they do not effect the asymptotic normality of SGA as discussed in the previous subsection. 
Then, we calculate the gradient of the adversary with each transformation and denote it as the augmentation-aware perturbation $\delta_i^{(t)}$:
\begin{equation} \label{eq:delta_t}
    \delta_i^{(t)} =\nabla_{x^{(t)}} \mathcal{L}(\texttt{AUG}_i(x_{adv}^{(t)}),y).
\end{equation}
Let $\mathcal{S}$ be a set of samples augmented in each iteration, i.e., $\mathcal{S} =\{\text{AUG}_i(x_{adv}^{(t)})\}_{i=1}^n$. Here $\mathcal{S}$ can be taken as a minibatch to form a stochastic gradient $\hat{\delta}_S$ which is a sum of contributions from all samples in $\mathcal{S}$. Hence, referring to the central limit theorem, $\hat{\delta}_S$ can be taken as a function of sample $z\in \mathcal{S}$,
\begin{equation} \label{eq:delta_exp}
    \hat{\delta}_S(z) \approx \delta(z) + \frac{1}{\sqrt{n}}\Delta \delta(z),
\end{equation}
where the gradient noises $\Delta \delta(z)$ at each update can be conveniently assumed following a Gaussian distribution with covariance $C(z)$,
\begin{equation} \label{eq:delta_norm}
    \begin{gathered}
        \Delta \delta(z) \sim \mathcal{N}(0,C(z)), \quad \hat{\delta}_S(z) \sim \mathcal{N}(\delta(z),\frac{1}{n}C(z)).
    \end{gathered}
\end{equation}
In this case, the expectation of the stochastic gradient is the full gradient, i.e.,
$\delta(z) = \mathbb{E}[\hat{\delta}_S(z)]$.
Hence, the prior distribution over adversarial perturbations is naturally a Gaussian one at each iteration.
Benefiting from the asymptotic normality of SGD~\cite{Mandt17}, we propose a novel method to approximate the posterior distribution of perturbations. Our main idea is to estimate the mean and covariance matrix of the stationary distribution of stochastic gradient ascent during the optimizing procedure. 
The iterative averaging is adopted to approximate the mean of adversarial perturbations using the sequence of stochastic gradients
\begin{equation}
    \bar{\delta}^{(t+1)} = \frac{(t \times n )\bar{\delta}^{(t)} +  \sum_{i=1}^n\delta_{i}^{(t)}}{(t+1)\times n}. \\
\end{equation}
Then, $\bar{\delta} = \bar{\delta}^{(T)}$, where $n$ is the augmented samples in the $t$-th iteration ($t=0,\cdots, T-1$) and $T$ is the total iterations. 

Let $\mathbf{D}$ be the deviation matrix with column $\mathbf{D}_j = \delta_i^{(t)} - \bar{\delta}^{(t+1)}$, where $j=t\times n+i$ and $i=1,\cdots, n$. Thus, the covariance matrix of the posterior distribution is estimated by considering all stochastic gradients in iterative processing,
\begin{equation}
    \sigma = \frac{\mathbf{D}\mathbf{D}^T}{n\times T-1}.
\end{equation}
Finally, the adversarial example can be iteratively obtained by (\ref{eq:algo_stochastic_bim}). Note that, ANDA allows to craft one or an unlimited number of adversarial examples for one input by sampling perturbations from the estimated distribution $\mathcal{N}\left(\bar{\delta}, \sigma \right)$.
The complete procedure for the proposed ANDA method is shown in Algorithm \ref{alg:example}. 
Following the empirical study in Section~\ref{sec:condi}, we visualized the perturbations $\delta_{i}^{(t)}$ of 10 iterates with multiple data augmentations (shown in Fig.~\ref{fig:Gaussians}~\textbf{b}), which validate our previous hypothesis.

\subsection{Multiple Implementations of ANDA}
\label{sec:multianda}
With the proposed ANDA, we find a solution to approximating the true posterior distribution centered at the the optimal $x^*_{adv}$  for the maximization problem introduced in (\ref{eq:algo_bim}). Nevertheless, only exploring the unimodal optimization space largely limits the the diversity of generated adversarial examples. Profiting the excellent generalization capability of Bayesian marginalization approximated with the deep ensemble strategy, we propose MultiANDA via multiple (e.g., K times) repeats of  ANDA by add random initial values on origin sample $x$. Especially, we average the adversarial perturbation $\bar{\delta}_k$ from each ANDA process, and obtain $\bar{\delta}_{mean} = \frac{1}{K} \sum_{k=0}^{K-1} \bar{\delta}_k$. This realizes the multimodal marginalization of this Gaussian mixture distribution.
Similarly, adversarial examples can be iteratively obtained by (\ref{eq:algo_bim}). MultiANDA also enables producing infinite adversarial examples with high effectiveness by sampling from each approximated Gaussian distribution $\mathcal{N}\left(\bar{\delta}_k, \sigma_k \right)$.
Remarkable improvements of the attacking ability, especially on the defense models are shown in Sections~\ref{subsec:ntmodels} and \ref{subsec:dmodels}. 

\section{Experiments}
\label{sec:exper}

In this section, we present the results of an extensive empirical study to validate the performance of the proposed methods.
We first specify the experimental settings in Section~\ref{subsec:setup}.
Then, the results of ANDA and MultiANDA compared with the state-of-the-art approaches on various types of DNNs with or without defenses are presented in Section~\ref{subsec:ntmodels} and \ref{subsec:dmodels}.
Section~\ref{subsec:dist} shows the attacking effectiveness of the multiple adversarial examples generated from each input via the proposed methods.
Due to page limitations, more detailed experiment settings, results and the ablation study are presented in Appendix.
\subsection{Experimental Setup}
\label{subsec:setup}
The datasets, models, and baseline methods employed in our experiments are detailed below.

\textbf{Datasets:} Following the prior work ~\cite{Dong18,Gao2020,Wang21b}, we implemented our experiments on ImageNet1k dataset \cite{imagenet1k} containing 1000 images, which are randomly drawn from the ImageNet dataset \cite{Rus15}.

\textbf{Models:} 
We considered two categories of target models for evaluation: seven \textit{normally trained models} and seven \textit{advanced defense models}. The first category includes Inception-v3 (Inc-v3)~\cite{SZE16}, Resnet-v2-50 (ResNet-50), Resnet-v2-101 (ResNet-101), Resnet-v2-152 (ResNet-152)~\cite{He16}, Inception-v4 (Inc-v4), Inception-ResNet-v2 (IncRes-v2)~\cite{SZE17} and VGG-19~\cite{Simonyan14vgg}, where Inc-v3, ResNet-50, IncRes-v2 and VGG-19 are also used as the white-box source models to generate adversarial examples. The defense models contain three adversarially trained models: Inc-v3$_{ens3}$, Inc-v3$_{ens4}$ and IncRes-v2$_{ens}$ ~\cite{Tramer17ensemble}; the  High-level representation Guided Denoiser (HGD) ~\cite{Liao2018hgd}; the Neural Representation Purifier (NRP) ~\cite{Naseer2020nrp}; the Randomized Smoothing (RS) ~\cite{Cohen19certified}; and the `Rand-3' submission in the NIPS 2017 defense competition (NIPS-r3).

\textbf{Baselines:} Three streams of transfer-based attack methods are used as baseline methods. The first stream focuses on the data augmentations including DIM~\cite{Xie19}, TIM~\cite{Dong19}, SIM~\cite{Wang21a} and a recent work Transferable Attack based on Integrated Gradients (TAIG)~\cite{Huang2022transferable}. For the second group, we choose the feature importance-aware (FIA)\cite{Wang21b}, which considers the middle-layer information of the source model and achieves the competitive performance.
The last stream focuses on the optimization enhanced methods, including BIM~\cite{Kurakin16}, its accelerated version: MI-FGSM~\cite{Dong18}, NI-FGSM~\cite{Lin20}, and the approaches using variance reduction strategy, VMI-FGSM, VNI-FGSM~\cite{Wang21a}.

\subsection{Attack Normally Trained Models}
\label{subsec:ntmodels}
We cross-validated our method by generating and testing the adversarial examples on the selected source (on rows) and target (on columns) networks (shown in Table~\ref{tab:normaltrained}). 
To be a fair comparison, our proposed methods generated one adversarial example for each input, aligned with the first output option of Algorithm~\ref{alg:example}, in the following experiments if not specified. From the Table~\ref{tab:normaltrained}, we observe that ANDA and MultiANDA significantly outperforms other methods in almost all cases, no matter in white-box or black-box settings.
As the numerical results demonstrate, they consistently defeat all other methods when taking IncRes-v2, ResNet-50 and VGG-19 as source models, and for Inc-v3, ANDA is competitive with the current SOTA methods, and MultiANDA always keep higher success rate.



\begin{figure}[ht]
    \centering
    \includegraphics[width=0.9\linewidth]{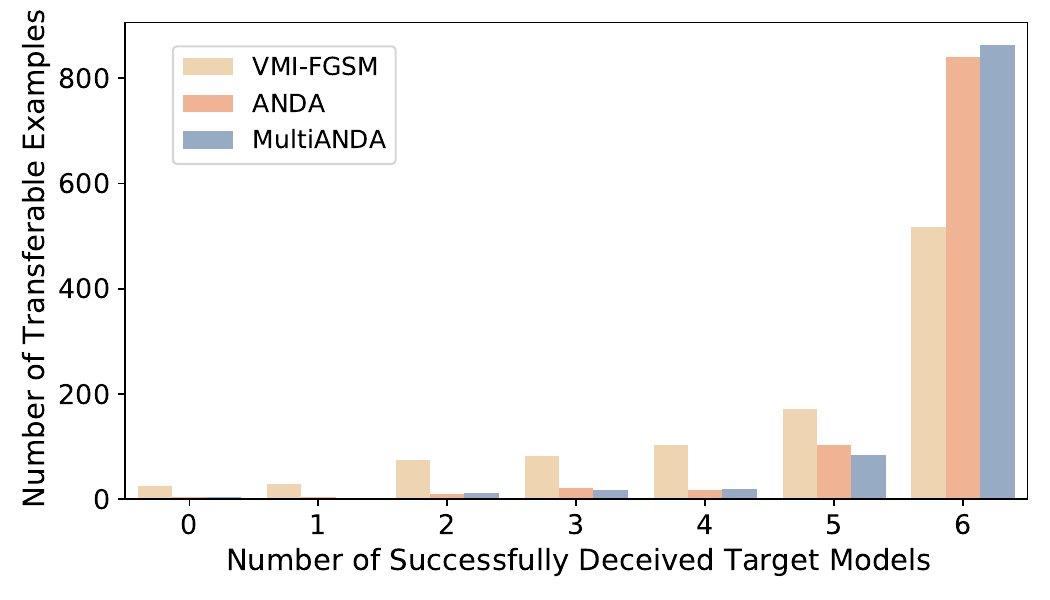}
    \caption{Most examples crafted by our proposed methods successfully deceived all 6 black-box models.}
    \label{fig:suc_normal}
\end{figure}

We further investigated the adversarial examples generated with ResNet-50, shown in Figure~\ref{fig:suc_normal}. The other six models were used as black-box targets for evaluating the performance of these examples. A higher number of successfully deceived target models implies more capable adversaries and therefore stronger attacks. The rightmost bars shows that ANDA, MultiANDA generated 840, 863/1000 adversaries that fooled all six targeted models, respectively, while 517/1000 are for VMI-FGSM. Naturally, the remaining sets of bars for this baseline method are higher, meaning that it deceived fewer targeted models.
Moreover, we visualized the generated examples and the perturbations by these three methods (see Figure~\ref{fig:vis_main}). The results show that the perturbations crafted by ANDA and MultiANDA focus more on the semantic areas of objects than VMI-FGSM, which dominates the decision of the prediction model. More visualization results are provided in Appendix.

\begin{table*}[!t]
    \footnotesize
    \centering
    \vspace{0.2cm}
    \setlength\tabcolsep{5pt}
    \begin{tabular}{c|cccccc|cccccc}
        \toprule
         \multirow{2}{*}{Attack} & \multicolumn{6}{c|}{Inc-v3 $\Longrightarrow$} & \multicolumn{6}{c}{ResNet-50 $\Longrightarrow$}                                                                              \\
                    & Inc-v3$_{ens3}$      & IncRes-v2$_{ens}$     & HGD              & NRP              & NIPS-r3            & Avg.                 & Inc-v3$_{ens3}$     & IncRes-v2$_{ens}$      & HGD               & NRP              & NIPS-r3            & Avg.              \\\hline
BIM                 & 11.1                 & 4.6                   & 3.7              & 13.4             & 4.8                & 7.5                  & 12.3                & 7.1                    & 7.4              & 13.7             & 8.1               & 9.7               \\
TIM                 & 27.5                 & 21.3                  & 16.9             & 22.8             & 21.1               & 21.9                 & 34.0                & 27.4                   & 24.2             & 31.1             & 30.6              & 29.5               \\
SIM                 & 18.1                 & 8.4                   & 8.6              & 15.2             & 10.8               & 12.2                 & 21.4                & 13.1                   & 15.2             & 16.1             & 15.1              & 16.2               \\
DIM                 & 13.1                 & 6.7                   & 5.8              & 12.8             & 8.6                & 9.4                  & 20.8                & 12.0                   & 16.9             & 15.7             & 15.3              & 16.1               \\
FIA                 & 37.4                 & 21.3                  & 11.6             & 23.5             & 29.2               & 24.6                 & 44.4                & 27.2                   & 30.0             & 25.9             & 35.3              & 32.6               \\
TAIG                & 38.0                 & 23.9                  & 22.8             & \textbf{29.6}    & 28.5               & \underline{28.6}     & 39.2                & 30.4                   & 29.7             & 29.6             & 34.4              & 32.7               \\
MI-FGSM             & 18.3                 & 9.0                   & 5.5              & 15.7             & 12.0               & 12.1                 & 26.3                & 15.5                   & 17.3             & 18.0             & 20.2              & 19.5               \\
NI-FGSM             & 18.6                 & 8.6                   & 6.2              & 15.3             & 12.2               & 12.2                 & 26.2                & 15.7                   & 17.5             & 18.9             & 19.8              & 19.6               \\
VMI-FGSM            & 36.9                 & 21.2                  & 19.1             & 24.7             & 27.7               & 25.9                 & 47.4                & 31.3                   & 37.7             & 27.1             & 38.7              & 36.4               \\
VNI-FGSM            & 36.4                 & 22.0                  & 18.9             & \underline{25.3} & 27.4               & 27.1                 & 46.5                & 30.4                   & 37.4             & 27.1             & 38.7              & 36.4               \\\cline{1-13}
ANDA                & \underline{44.4}     & \underline{25.9}      & \underline{36.5} & 23.2             & \underline{37.0}   & 20.3                 & \underline{71.4}    & \underline{50.7}       & \underline{72.4} & \underline{32.9} & \underline{67.4}  & \underline{41.6 }  \\
MultiANDA           & \textbf{54.4}        & \textbf{36.7}         & \textbf{52.8}    & 24.3             & \textbf{46.9}      & \textbf{29.9}        & \textbf{79.7}       & \textbf{64.8}          & \textbf{82.3}    & \textbf{34.4}    & \textbf{76.3}     & \textbf{50.1 }     \\
        \hline            
        \multirow{2}{*}{Attack} & \multicolumn{6}{c|}{IncRes-v2 $\Longrightarrow$} & \multicolumn{5}{c}{VGG-19 $\Longrightarrow$}                                                                                          \\
                    & Inc-v3$_{ens3}$      & IncRes-v2$_{ens}$   & HGD              & NRP              & NIPS-r3           & Avg.                 & Inc-v3$_{ens3}$         & IncRes-v2$_{ens}$      & HGD               & NRP              & NIPS-r3            & Avg.          \\\hline
BIM                 & 11.1                 & 7.0                 & 5.1              & 13.0             & 6.0               & 8.4                  & 9.6                     & 4.4                    & 3.9              & 13.4             & 4.6               & 7.2             \\
TIM                 & 27.9                 & 21.4                & 18.4             & 23.9             & 23.8              & 23.1                 & \underline{23.0}        & 27.4                   & 15.0             & 31.1             & 30.6              & \textbf{25.4}             \\
SIM                 & 23.8                 & 16.4                & 15.9             & 18.3             & 18.1              & 18.5                 & 11.4                    & 5.6                    & 8.7              & 13.4             & 7.4               & 9.3             \\
DIM                 & 16.2                 & 10.1                & 11.7             & 14.9             & 12.3              & 13.0                 & 10.7                    & 4.7                    & 5.6              & 13.6             & 5.9               & 8.1             \\
FIA                 & 48.9                 & 34.7                & 24.2             & \underline{30.8} & 42.8              & 36.3                 & 13.3                    & 8.3                    & 7.1              & 14.6             & 9.8               & 10.6             \\
TAIG                & 49.0                 & 41.2                & 12.2             & 16.5             & 13.3              & 26.4                 & 17.7                    & 10.2                   & \textbf{36.9}    & \textbf{34.1}    & \textbf{41.9}     & 28.2             \\
MI-FGSM             & 22.0                 & 13.3                & 13.4             & 16.4             & 17.0              & 16.4                 & 13.0                    & 6.9                    & 7.2              & 15.5             & 8.9               & 10.3             \\
NI-FGSM             & 21.6                 & 13.8                & 13.4             & 15.5             & 16.6              & 16.2                 & 13.9                    & 6.8                    & 7.2              & 14.6             & 9.3               & 10.4             \\
VMI-FGSM            & 49.2                 & 38.8                & 36.0             & 25.2             & 39.2              & 37.7                 & 21.9                    & 11.6                   & 15.9             & 18.4             & 15.9              & 16.7             \\
VNI-FGSM            & 49.9                 & 37.7                & 35.4             & 25.6             & 39.3              & \underline{38.4}     & 21.6                    & \underline{11.9}       & 16.8             & \underline{18.6} & 15.7              & \underline{22.9}             \\\cline{1-13}
ANDA                & \underline{63.3}     & \underline{47.3}    & \underline{57.8} & 28.5             & \underline{57.9}  & 32.6                 & 20.9                    & 10.8                   & 22.2             & 15.4             & 16.2              & 3.7             \\
MultiANDA           & \textbf{70.3}        & \textbf{61.9}       & \textbf{69.9}    & \textbf{31.2}    & \textbf{67.8}     & \textbf{41.8}        & \textbf{24.6}           & \textbf{13.7}          & \underline{31.2} & 16.2             & \underline{21.3}  & 7.9           \\
                   
        \bottomrule
    \end{tabular}
   \caption{\label{tab:defense}
       The success rates (\%) of the proposed and baseline attacks on five defense models. The best/second results are shown in bold/underlined. Due to the space limit, performance on five target models are presented here. See full results on seven targets in Appendix.
   }
\end{table*}

\subsection{Attack Defense Models}
\label{subsec:dmodels}

We evaluate our proposed method targeting on the selected advanced defenses models.
As shown in Table~\ref{tab:defense}, the success rates of our proposed methods consistently improve over these various defense models in most cases. 
They provides a significant performance increase in general compared with the other attacks, especially reaches the overall optimal success rates with the source model ResNet-50. An intriguing finding is that using ResNet-50 as the source model always yields the best attacking performance. This aligns with the conclusion drawn by Wu \etal~\cite{Wu2020} that the skip connections in ResNet-like neural networks enhance black-box attack transferability.

Notably, the fluctuating performance and the generally reduced effectiveness of all baseline methods can be observed with the source model VGG-19. The average maximum attack success rate in this context is only 28.3\%, suggesting that the feature extraction capabilities of VGG-19 are not optimal, thereby leading to diminished transferability for black-box attacks. This observation highlights the critical role of source model selection in maximizing the transferability of adversarial attacks. Hence, the choice of source models  and enhancing their capability for generalized feature extraction become promising directions for the future research. 

Similarly, we visualized the generated examples and the perturbations by VMI-FGSM, ANDA and MultiANDA that fool the defense models. The perturbed area by our proposed methods concentrates more on the informative region of images than the baseline method. Details are illustrated in Appendix. These compelling results of ANDA and MultiANDA on defenses valid their overall generalization ability in generating transferable and diverse adversaries.

\begin{figure}[ht]
    \centering
    \includegraphics[width=\linewidth]{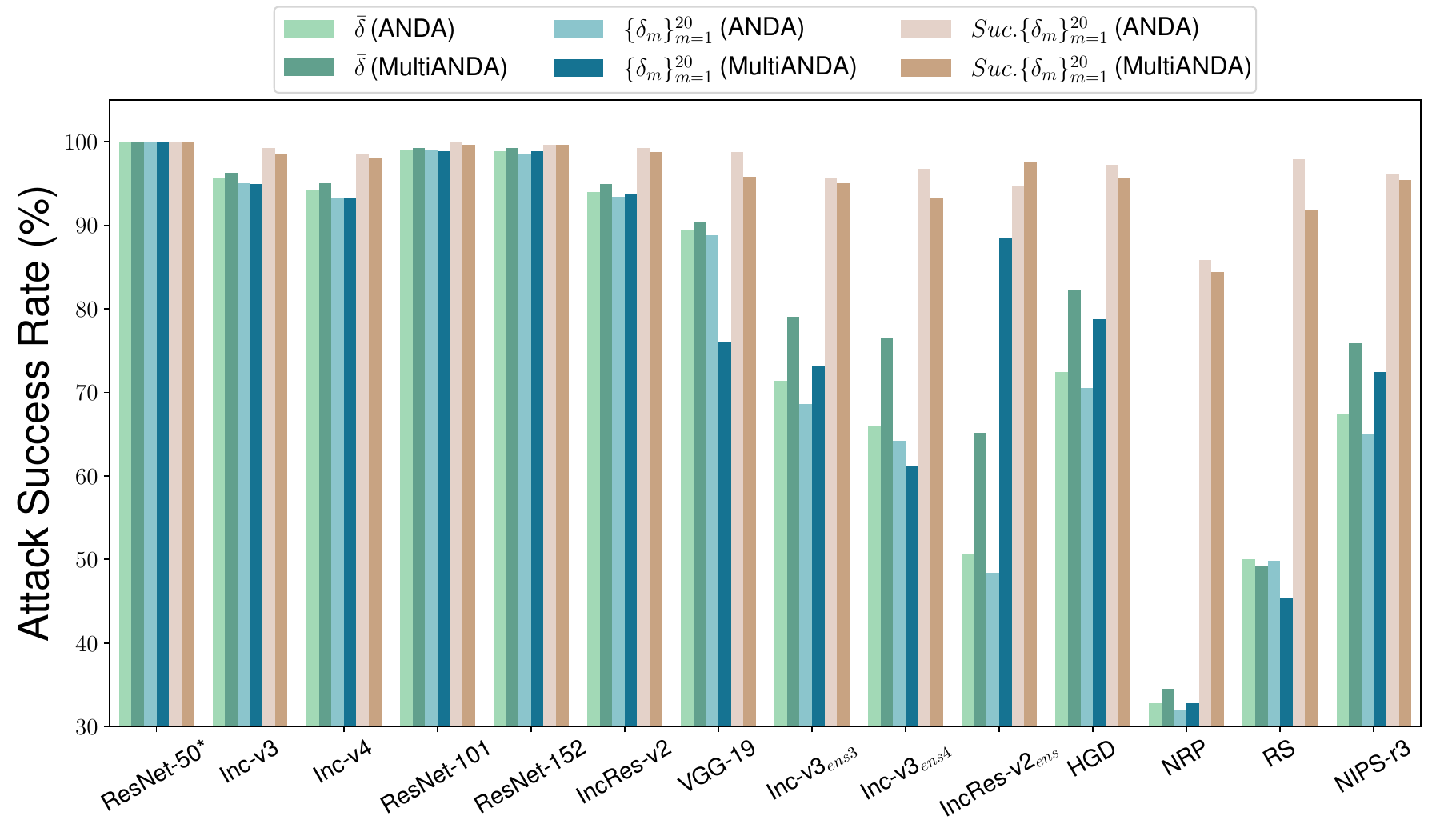}
    \vspace{-0.3cm}
    \caption{Attack success rates of sampled adversaries with single  $\{\bar{\delta}\}$, multiple $\{\delta_{m}\}_{m=1}^{20}$  and the succeeded $Suc. \{\delta_{m}\}_{m=1}^{20}$ generated by ANDA and MultiANDA. $\{\delta_{m}\}_{m=1}^{20}$ means the 20 adversarial examples sampled from the learned perturbation distribution for each input image; whereas $Suc. \{\delta_{m}\}_{m=1}^{20}$ are the 20 sampled adversarial examples for the input image whose corresponding single adversarial example has succeeded in the attack.
    }\label{fig:success_rates_plots}
\end{figure}

\subsection{Attack Performance of Sampled Adversaries}
\label{subsec:dist}

The analysis presented above shows the effectiveness of the single optimum adversarial example per input generated, denoted as $\bar{\delta}$, referring to \textit{Step 6 (a)} shown in Algorithm~\ref{alg:example}. Besides, MultiANDA/ANDA allows learning the  distribution of the perturbations $\mathcal{N}\left(\bar{\delta}, \sigma \right)$ via \textit{Step 6 (b)}, and generate any number of adversarial examples for each input by sampling $M$ adversaries, denoted as $\{\delta_{m}\}_{m=1}^{M}$.
To verify the attack performance of these sampled adversaries, we drawn 20 ($M=20$) adversarial examples for each input image, i.e., $\{\delta_{m}\}_{m=1}^{20}$. Meanwhile, we analyzed the adversaries sampled for the input images whose corresponding adversaries $\bar{\delta}$ succeeded in the attack, denoted as $Suc. \{\delta_{m}\}_{m=1}^{20}$.
Experiment results show in Figure~\ref{fig:success_rates_plots} that
the sampled adversaries have competitive attack success rates compared with the single adversary generated with $\bar{\delta}$, even on the defense models. This confirms the well approximation of the perturbation distributions and the great potential of our method for adversarial training or defense model designing. 

\section{Conclusions and Perspectives}
\label{sec:concl}
In this work, we present the Multiple Asymptotically Normal Distribution Attacks (MultiANDA) to generate transferable adversarial examples. Our primary goal is to efficiently approximate the posterior distribution of adversarial perturbations to achieve better generalization across unknown deep learning models. We leverage the statistical information of gradients along the optimization trajectory to estimate the stationary distribution of the perturbations. With this learned distribution's aid, MultiANDA can significantly improve the transferability of adversarial examples to black-box target models. Extensive experiments demonstrate that MultiANDA outperforms popular and recently proposed black-box and transfer-based methods, further highlighting the insufficiency of current defense techniques. Notably, the computational overhead of both ANDA and MultiANDA, largely depending on the number of batch samples during augmentation, is on par with similar baseline methods (refer to Appendix for the computational overhead analysis). Compared to the potential risks these algorithms might pose if misused, their computational cost is relatively negligible. We advocate for the development of advanced deep learning models which align better with human visual preferences, thereby avoiding the fundamental defects present in modern models.


\section*{Acknowledgements}
This work was supported by the National Natural Science Foundation of China under Grant 62176020, the Joint Foundation of the Ministry of Education (8091B042235), the Beijing Natural Science Foundation under Grant L211016, the Talent Fund of Beijing Jiaotong University under Grant 2023XKRC041, and the Fundamental Research Funds for the Central Universities (Science and technology leading talent team project) under Grant 2022JBXT003. We also thank Huafeng Liu and Zhekun Liu for their helpful advice and support. 

{
    \small
    \bibliographystyle{ieeenat_fullname}
    \bibliography{main}
}

\clearpage
\maketitlesupplementary
\appendix

\section{Implementation Details}
\label{appen:imple}

\subsection{Algorithm for Multiple Asymptotically Normal Distribution Attack}

In Section~3,  we propose the Multiple Asymptotically Normal Distribution Attacks (MultiANDA), a novel method that explicitly characterizes perturbations inferred from a learned distribution. Specifically, in Section~3.2, we first elaborate on the procedure of single ANDA, which approximates the posterior distribution over the adversarial perturbations by leveraging the asymptotic normality property of stochastic gradient ascent (SGA). 
Subsequently, in Section~3.3, we apply an ensemble strategy on ANDA to estimate a mixture of Gaussian distributions, enhancing exploration of the potential optimization space. This leads to further improved generalization performance of the attacks, facilitating the generation of more diverse and transferable adversarial examples. The detailed implementation of MultiANDA is shown in Algorithm~\ref{alg:multianda}.

\begin{algorithm}\small
    \caption{Multiple Asymptotically Normal Distribution Attacks}
    \setlength{\baselineskip}{9pt}
    \label{alg:multianda}




     \textbf{Input:} $x$: clean image $x \in \mathbb{R}^{d}$; $y$: ground-truth label; $f$: pre-trained source model;
    
    \textbf{Parameters:} $T$: \# iterations; $\epsilon$: perturbation magnitude; $n$: \# batch samples for augmentation; $M$: \# sampling examples; $K$: \# ensemble ANDAs; $\gamma$: the radius of uniform noise;

    \textbf{Output:} $x_{adv}$: Adversarial examples;

    \begin{algorithmic}[tb]
        \STATE \textbf{Initialize } $\alpha \leftarrow \epsilon / T $, $x_0 \leftarrow x$
        \FOR{$k=0$ {\bfseries to} $K-1$}
        \STATE{Initialize the uniform noise in parallel} 
        \begin{equation*}
            \begin{aligned}
                 & x_{k}^{(0)} = x^{(0)} + u_{k},  u_k \sim \mathcal{U}(-\gamma, \gamma) \\
            \end{aligned}
        \end{equation*}

        \STATE Run $K$ ANDA loops independently and in parallel (Refer ANDA to Algorithm~1 in Section~3 )
        \begin{equation*}
            \begin{aligned}
                 & x_{k}^{(T-1)}, \bar{\delta}_k^{(T)}, \sigma_k^{(T)} = \text{ANDA}(x_k^{(0)}, y, f, T, M, \epsilon, n) \\
            \end{aligned}
        \end{equation*}
        \ENDFOR
        \STATE Average $K$ ANDA outputs of penultimate iterations:
        \begin{equation*}
            \begin{aligned}
                \overline{x^{(T-1)}} = \frac{1}{K} \sum_{k=0}^{K-1} x_{k}^{(T-1)}
            \end{aligned}
        \end{equation*}
        \STATE {Output option (a): Craft one $x_{adv}$ using the mean perturbation $\bar{\delta}_{mean}$\\
            \begin{equation*}
                \begin{aligned}
                     & \bar{\delta}_{mean} = \frac{1}{K} \sum_{k=0}^{K-1} \bar{\delta}_k^{(T)}                                           \\
                     & x_{adv} = \texttt{Clip}_{x, \epsilon} \{ \overline{x^{(T-1)}} + \alpha \cdot \texttt{Sign}(\bar{\delta}_{mean})\} \\
                \end{aligned}
            \end{equation*}
            Output option (b): Craft M $x_{adv}$ by sampling from the learned perturbation distribution $\mathcal{N}(\bar{\delta}_k,\sigma_k)$ $(k = 0, \dots, K-1)$
            \begin{equation*}
                \begin{aligned}
                     & \{\bar{\delta}_{m} = \frac{1}{K}\sum_{k=0}^{K-1} \delta_{k,m} , \quad \delta_{k,m}  \sim \mathcal{N}\left(\bar{\delta}_k^{(T)}, \sigma_k^{(T)} \right) \}_{m=1}^{M} \\
                     & \{x_{adv}^{m}\}_{m=1}^{M} = \{\texttt{Clip}_{x, \epsilon} \{ \overline{x^{(T-1)}} + \alpha \cdot \texttt{Sign}(\bar{\delta}_{m})\}\}_{m=1}^{M}
                \end{aligned}
            \end{equation*}

        }

    \end{algorithmic}
\end{algorithm}


\subsection{Datasets and Models}
In this subsection, the datasets and models employed in our experiments are detailed. In addition to the evaluation experiments introduced in Section 4, we provide more results on the ViT-based~\cite{ViT, DeiT, PiT} and CLIP~\cite{Rad21} models in this supplementary material, which are used to illustrate the generalizability of our proposed methods to various model architectures and the performance on the cross-domain task.

\textbf{Datasets}: The dataset is ImageNet-1k, which is identical to the one described in Section~4.1.

\textbf{Models}: We considered two categories of target models for evaluation: seven \textit{normally trained models} and seven \textit{advanced defense models}. The first category includes Inception-v3 (Inc-v3)~\cite{SZE16}, Resnet-v2-50 (ResNet-50), Resnet-v2-101 (ResNet-101), Resnet-v2-152 (ResNet-152)~\cite{He16}, Inception-v4 (Inc-v4), Inception-ResNet-v2 (IncRes-v2)~\cite{SZE17} and VGG-19~\cite{Simonyan14vgg}, where Inc-v3, ResNet-50, IncRes-v2 and VGG-19 are also used as the white-box source models to generate adversarial examples. The defense models contain three adversarially trained models: Inc-v3$_{ens3}$, Inc-v3$_{ens4}$ and IncRes-v2$_{ens}$ ~\cite{Tramer17ensemble}; the  High-level representation Guided Denoiser (HGD) ~\cite{Liao2018hgd}; the Neural Representation Purifier (NRP) ~\cite{Naseer2020nrp}; the Randomized Smoothing (RS) ~\cite{Cohen19certified}; and the `Rand-3' submission in the NIPS 2017 defense competition (NIPS-r3).

We aim to utilize ANDA/MultiAND to thoroughly test the defense effectiveness of various strategies in recently proposed defense approaches. The defense strategies and the implementation details of these selected \textit{advanced defense models} are presented as follows:

\begin{itemize}
    \item Three adversarially trained models, namely Inc-v3$_{ens3}$ , Inc-v3$_{ens4}$ and IncRes-v2$_{ens}$ ~\cite{Tramer17ensemble};
    \item High-level representation guided denoiser (HGD, rank-1 submission in the NIPS 2017 defense competition) ~\cite{Liao2018hgd};
    \item Rand-3 submission in the NIPS 2017 defense competition (NIPS-r3) \footnote{\url{https://github.com/anlthms/nips-2017/tree/master/mmd}};
    \item Randomized Smoothing (RS) ~\cite{Cohen19certified}
    \item Neural Representation Purifier (NRP) ~\cite{Naseer2020nrp}
\end{itemize}

We adopted the implementation of NRP as described in \cite{Wang21a}, and sourced the remaining defense methods from their official implementations. For all defense methods, we utilized Inc-v3$_{ens3}$ as the backbone architecture.

For the ViT series of models, four mainstream models: ViT-L/16 \cite{ViT}, DeiT3-B/16 \cite{DeiT}, Swin-B/4 \cite{Swin} and PiT-B \cite{PiT} are selected. For the evaluation against the cross-domain task, we choose the renowned multimodal model CLIP~\cite{Rad21}.


%

\subsection{Hyper-Parameters}
\label{sec:hp}
The basic attack settings were consistent with the work \cite{Wang21a,Dong18}: the maximum perturbation of $\epsilon = 16$, and the number of iterations $T=10$ and step size $\alpha = \epsilon / T = 1.6$. We carefully tuned the specific hyper-parameters for each baseline (e.g., the sampling number in VMI-FGSM, the ensemble number in FIA, etc.), and \textbf{the best results were recorded. }

For ANDA, we set the number of augmented samples as 25 (i.e., $n=25$), and the image augmenting parameter $\texttt{Augmax}$ (refer detailed explanation in Section~\ref{sec:augmax}) as $0.3$ if not specified.
For MultiANDA, the default number of ANDA components is $5$ (i.e., $K=5$).
To ensure the random starts of each MultiANDA component, we added small uniform noises $u$ to the original sample $x$, where $u \sim \mathcal{U}(-\gamma, \gamma)$, as shown in Algorithm~\ref{alg:multianda}. To balance the diversity among the ensemble components and the original semantic information of inputs, we set the radius of uniform noise $\gamma=\frac{0.5}{255}$. Regarding the adopted baselines, we employed the following corresponding hyper-parameters:
\begin{itemize}
    \item For MI-FGSM, NI-FGSM and their variants, we set the decay factor $\mu = 1.0$. For VMI-FGSM~\cite{Wang21a}, $n=20$ and $\beta=1.5$.
    \item For DIM~\cite{Xie19}, the transformation probability was set to $0.5$. For SIM~\cite{Wang21a}, the number of scale copies is 5. For TIM~\cite{Dong19}, we tuned the Gaussian kernel size in $\{15 \times 15$, $7 \times 7\}$ with the same standard deviation $\sigma = 3$ by referring their official code\footnote{\url{https://github.com/dongyp13/Translation-Invariant-Attacks}}.
    \item For FIA \cite{Wang21b}, we followed the official settings in the corresponding paper.
    \item For TAIG \cite{Huang2022transferable}, we chose TAIG-S as the baseline. To ensure a fair comparison in the single form, we adhered to our basic settings ($\epsilon = 16$, $T=10$, and the number of turning points $E = 20$). For a detailed comparison showcasing the best performance of TAIG, refer to Section~\ref{sec:taig}.

\end{itemize}

\section{Additional Experimental Results}
\label{appen:DER-NDA}

\subsection{Full evaluation results}

Taking advantage of the extensive space in this appendix, we present the complete experimental results for the baseline methods and the proposed ANDA/MultiAND. These encompass all seven normally trained models and seven advanced defense models, as detailed in Table~\ref{tab:normaltrained_compl} and Table~\ref{tab:defense_compl}, respectively. Notably, additional experiments involving Inc-v4, ResNet-101, and Inc-v3$_{ens4}$—models with structures similar to Inc-v3, ResNet-50 and Inc-v3$_{ens3}$—further validate the effectiveness of our methods. An intriguing finding is that using ResNet-50 as the source model yields the best attacking performance. This aligns with the conclusion drawn by Wu \etal~\cite{Wu2020}. They empirically studied this observation and disclosed the skip connection provided the actual contribution to this enhancement. Wu \etal~\cite{Wu2020} and Zhu \etal~\cite{Zhu21} further boost the transferability by exploiting this finding. Moreover, the comparatively modest results on RS models have sparked our interest for future research. 

\begin{figure}[ht]
    \centering
    \includegraphics[width=0.95\linewidth]{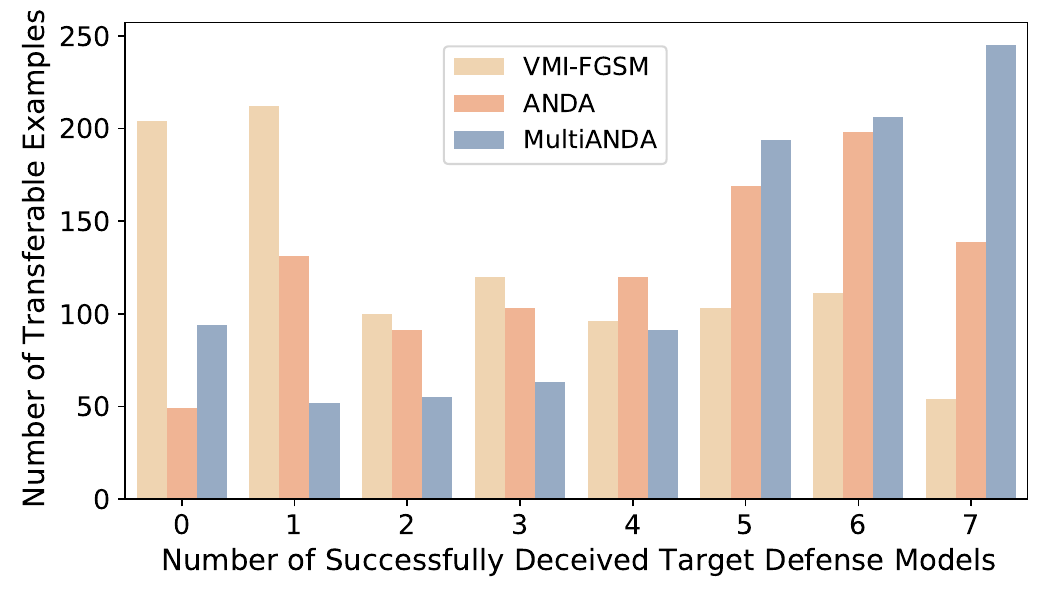}
    \caption{Most examples crafted by our proposed methods successfully deceived more than 5 defense models.}\label{fig:suc_on_model_defense}
\end{figure}

We conducted a similar statistical analysis, as shown in Figure~3 in the main body, on defense models to determine the numbers of black-box target models that each generated example can deceive. The source model remains ResNet-50. Figure~\ref{fig:suc_on_model_defense} shows 15\% and 25\% of the examples generated by ANDA and MultiANDA managed to fool all seven models (as seen in the rightmost bars), respectively. In contrast, only 5\% examples crafted by VMI-FGSM achieved this. In addition, the failure rate of VMI-FGSM across all seven models stands at 20\%, compared to only 5\% for ANDA. These results indicate that the generated adversaries by ANDA and MultiANDA are more diverse and have better transferability than those by VMI-FGSM.

\begin{table*}[h]
    \scriptsize
    \centering
    \begin{tabular}{c|c|rrrrrrr}
        \toprule
        \multirow{2}{*}{} & \multirow{2}{*}{Attack} & \multicolumn{7}{c}{Target model}                                                                                                                      \\\cline{3-9}
                          &                         & Inc-v3                           & Inc-v4           & ResNet-50         & ResNet-101       & ResNet-152       & IncRes-v2         & VGG-19            \\
        \midrule

        \multirow{10}{*}{\rotatebox[origin=c]{90}{IncRes-v2}}
                          & BIM                     & 36.3                             & 28.7             & 25.6              & 22.1             & 20.6             & 99.3*             & 37.8              \\
                          & TIM                     & 43.4                             & 35.3             & 36.5              & 33.2             & 32.0             & 36.0*             & 66.2              \\
                          & SIM                     & 58.9                             & 50.9             & 47.8              & 41.7             & 41.4             & \underline{99.6*} & 49.8              \\
                          & DIM                     & 54.6                             & 51.1             & 41.4              & 37.2             & 36.6             & 98.2*             & 50.0              \\
                          & FIA                     & 82.2                             & 78.3             & 75.3              & 75.2             & 72.4             & 89.2*             & 80.7              \\
                          & TAIG                    & 73.9                             & 69.1             & 63.4              & 60.5             & 58.4             & 95.0*             & 57.4              \\
                          & NI-FGSM                 & 61.9                             & 52.5             & 49.5              & 47.5             & 44.7             & 99.2*             & 64.7              \\
                          & MI-FGSM                 & 60.3                             & 53.4             & 49.3              & 45.6             & 43.0             & 98.8*             & 64.6              \\
                          & VMI-FGSM                & 81.1                             & 76.9             & 69.6              & 69.0             & 66.4             & 99.3*             & 73.5              \\
                          & VNI-FGSM                & 80.9                             & 77.2             & 70.0              & 68.6             & 65.8             & 99.4*             & 73.8              \\\cline{2-9}
                          & ANDA                    & \underline{93.0}                 & \underline{90.5} & \underline{86.4}  & \underline{83.9} & \underline{83.7} & \textbf{99.8*}    & \underline{82.8}  \\
                          & MultiANDA               & \textbf{93.9}                    & \textbf{92.1}    & \textbf{87.1}     & \textbf{86.1}    & \textbf{85.6}    & \textbf{99.8*}    & \textbf{84.3}     \\
        \hline
        \multirow{10}{*}{\rotatebox[origin=c]{90}{ResNet-50}}
                          & BIM                     & 33.2                             & 26.6             & 99.7*             & 70.5             & 63.9             & 20.9              & 43.4              \\
                          & TIM                     & 47.3                             & 41.3             & 77.0*             & 49.4             & 47.6             & 30.8              & 68.8              \\
                          & SIM                     & 48.6                             & 40.8             & \textbf{100.0*}   & 87.5             & 83.7             & 35.3              & 50.9              \\
                          & DIM                     & 61.9                             & 55.4             & \underline{99.9*} & 88.0             & 83.6             & 47.5              & 61.1              \\
                          & FIA                     & 86.2                             & 83.9             & 99.6*             & 95.3             & 94.5             & 80.4              & 88.3              \\
                          & TAIG                    & 62.4                             & 55.8             & \textbf{100.0*}   & 89.9             & 86.1             & 51.9              & 58.7              \\
                          & NI-FGSM                 & 59.6                             & 52.7             & 99.7*             & 88.0             & 85.4             & 48.1              & 67.0              \\
                          & MI-FGSM                 & 58.5                             & 52.6             & 99.7*             & 88.2             & 85.9             & 48.6              & 67.4              \\
                          & VMI-FGSM                & 75.3                             & 69.9             & \underline{99.9*} & 95.5             & 93.4             & 68.3              & 76.4              \\
                          & VNI-FGSM                & 75.4                             & 69.1             & 99.8*             & 95.0             & 92.9             & 67.9              & 75.6              \\\cline{2-9}
                          & ANDA                    & \underline{95.6}                 & \underline{94.3} & \textbf{100.0*}   & \underline{99.0} & \underline{98.9} & \underline{94.0}  & \underline{89.5}  \\
                          & MultiANDA               & \textbf{96.5}                    & \textbf{94.9}    & \textbf{100.0*}   & \textbf{99.2}    & \textbf{99.2}    & \textbf{95.0}     & \textbf{90.1}     \\
        \hline
        \multirow{10}{*}{\rotatebox[origin=c]{90}{VGG-19}}
                          & BIM                     & 23.5                             & 23.9             & 18.7              & 17.4             & 13.7             & 9.9               & \underline{99.9*} \\
                          & TIM                     & 41.5                             & 35.3             & 34.8              & 31.5             & 30.2             & 23.6              & \textbf{100.0*}   \\
                          & SIM                     & 37.7                             & 43.3             & 34.4              & 29.2             & 25.2             & 23.6              & \textbf{100.0*}   \\
                          & DIM                     & 31.7                             & 35.8             & 26.4              & 21.9             & 19.2             & 16.4              & \underline{99.9*} \\
                          & FIA                     & 57.4                             & 61.3             & 50.7              & 44.9             & 40.9             & 42.7              & \textbf{100.0*}   \\
                          & TAIG                    & 48.8                             & 52.6             & 43.6              & 35.8             & 34.7             & 33.9              & \textbf{100.0*}   \\
                          & NI-FGSM                 & 43.6                             & 48.7             & 39.5              & 34.8             & 29.2             & 30.4              & \underline{99.9*} \\
                          & MI-FGSM                 & 44.7                             & 48.1             & 39.4              & 34.7             & 28.9             & 30.8              & \underline{99.9*} \\
                          & VMI-FGSM                & 62.7                             & 65.3             & 56.7              & 49.3             & 46.5             & 48.6              & \textbf{100.0*}   \\
                          & VNI-FGSM                & 63.2                             & 65.6             & 56.7              & 50.4             & 46.6             & 48.8              & \textbf{100.0*}   \\\cline{2-9}
                          & ANDA                    & \underline{74.4}                 & \underline{80.6} & \underline{64.1}  & \underline{59.8} & \underline{56.4} & \underline{61.5}  & \textbf{100.0*}   \\
                          & MultiANDA               & \textbf{75.4}                    & \textbf{82.1}    & \textbf{66.1}     & \textbf{61.5}    & \textbf{58.6}    & \textbf{63.5}     & \textbf{100.0*}   \\
        \hline
        \multirow{10}{*}{\rotatebox[origin=c]{90}{Inc-v3}}
                          & BIM                     & \textbf{100.0*}                  & 24.1             & 20.3              & 16.4             & 15.7             & 15.6              & 34.3              \\
                          & TIM                     & 64.3*                            & 35.9             & 35.9              & 31.6             & 30.6             & 25.4              & 70.4              \\
                          & SIM                     & \textbf{100.0*}                  & 43.3             & 38.2              & 32.9             & 31.1             & 35.9              & 42.2              \\
                          & DIM                     & \textbf{100.0*}                  & 42.5             & 31.7              & 26.9             & 25.5             & 31.4              & 45.5              \\
                          & FIA                     & 98.3*                            & 85.4             & \underline{78.4}  & \underline{76.1} & \underline{75.3} & 81.2              & \textbf{84.5}     \\
                          & TAIG                    & \underline{99.7*}                & 59.6             & 53.3              & 47.8             & 45.9             & 56.7              & 54.2              \\
                          & NI-FGSM                 & \textbf{100.0*}                  & 44.9             & 40.0              & 37.3             & 35.2             & 39.9              & 56.9              \\
                          & MI-FGSM                 & \textbf{100.0*}                  & 45.1             & 40.2              & 36.2             & 35.1             & 40.3              & 57.1              \\
                          & VMI-FGSM                & \textbf{100.0*}                  & 71.7             & 63.0              & 58.9             & 59.3             & 68.6              & 70.3              \\
                          & VNI-FGSM                & \textbf{100.0*}                  & 72.6             & 62.4              & 58.3             & 58.7             & 67.7              & 69.7              \\\cline{2-9}
                          & ANDA                    & \textbf{100.0*}                  & \underline{85.7} & 76.1              & 74.1             & 72.8             & \underline{82.3}  & 77.0              \\
                          & MultiANDA               & \textbf{100.0*}                  & \textbf{88.2}    & \textbf{79.2}     & \textbf{77.3}    & \textbf{76.0}    & \textbf{84.5}     & \underline{78.8}  \\

        \bottomrule
    \end{tabular}
    \caption{\label{tab:normaltrained_compl}
        Attack success rates (\%) of the proposed and baseline attacks against seven normally trained models. Sign * indicates the results against white-box models. The best/second results are shown in bold/underlined. 
    }
\end{table*}

\begin{table*}[h]
    \scriptsize
    \centering
    \begin{tabular}{c|c|ccccccc}
        \toprule
        \multirow{2}{*}{} & \multirow{2}{*}{Attack} & \multicolumn{7}{c}{Target model}                                                                                                                    \\\cline{3-9}
                          &                         & Inc-v3$_{ens3}$                  & Inc-v3$_{ens4}$  & IncRes-v2$_{ens}$ & HGD              & RS               & NRP              & NIPS-r3          \\
        \hline
        \multirow{10}{*}{\rotatebox[origin=c]{90}{IncRes-v2}}
                          & BIM                     & 11.1                             & 12.3             & 7.0               & 5.1              & 17.7             & 13.0             & 6.0              \\
                          & TIM                     & 27.9                             & 26.6             & 21.4              & 18.4             & \textbf{54.8}    & 23.9             & 23.8             \\
                          & SIM                     & 23.8                             & 23.0             & 16.4              & 15.9             & 22.6             & 18.3             & 18.1             \\
                          & DIM                     & 16.2                             & 14.7             & 10.1              & 11.7             & 19.5             & 14.9             & 12.3             \\
                          & FIA                     & 48.9                             & 45.0             & 34.7              & 24.2             & \underline{46.1} & \underline{30.8} & 42.8             \\
                          & TAIG                    & 49.0                             & 46.8             & 41.2              & 12.2             & 26.0             & 16.5             & 13.3             \\
                          & MI-FGSM                 & 22.0                             & 21.6             & 13.3              & 13.4             & 28.8             & 16.4             & 17.0             \\
                          & NI-FGSM                 & 21.6                             & 21.9             & 13.8              & 13.4             & 28.6             & 15.5             & 16.6             \\
                          & VMI-FGSM                & 49.2                             & 42.9             & 38.8              & 36.0             & 36.7             & 25.2             & 39.2             \\
                          & VNI-FGSM                & 49.9                             & 44.1             & 37.7              & 35.4             & 36.5             & 25.6             & 39.3             \\\cline{2-9}
                          & ANDA                    & \underline{63.3}                 & \underline{58.2} & \underline{47.3}  & \underline{57.8} & 43.9             & 28.5             & \underline{57.9} \\
                          & MultiANDA               & \textbf{70.3}                    & \textbf{66.3}    & \textbf{61.9}     & \textbf{69.9}    & 41.5             & \textbf{31.2}    & \textbf{67.8}    \\
        \hline
        \multirow{10}{*}{\rotatebox[origin=c]{90}{ResNet-50}}
                          & BIM                     & 12.3                             & 13.6             & 7.1               & 7.4              & 19.0             & 13.7             & 8.1              \\
                          & TIM                     & 34.0                             & 33.9             & 27.4              & 24.2             & \textbf{58.2}    & 31.1             & 30.6             \\
                          & SIM                     & 21.4                             & 20.9             & 13.1              & 15.2             & 23.1             & 16.1             & 15.1             \\
                          & DIM                     & 20.8                             & 21.4             & 12.0              & 16.9             & 22.5             & 15.7             & 15.3             \\
                          & FIA                     & 44.4                             & 39.5             & 27.2              & 30.0             & 43.5             & 25.9             & 35.3             \\
                          & TAIG                    & 39.2                             & 40.8             & 30.4              & 29.7             & 38.1             & 29.6             & 34.4             \\
                          & MI-FGSM                 & 26.3                             & 23.9             & 15.5              & 17.3             & 32.2             & 18.0             & 20.2             \\
                          & NI-FGSM                 & 26.2                             & 25.2             & 15.7              & 17.5             & 31.9             & 18.9             & 19.8             \\
                          & VMI-FGSM                & 47.4                             & 45.3             & 31.3              & 37.7             & 43.4             & 27.1             & 38.7             \\
                          & VNI-FGSM                & 46.5                             & 45.1             & 30.4              & 37.4             & 43.3             & 27.1             & 38.7             \\\cline{2-9}
                          & ANDA                    & \underline{71.4}                 & \underline{65.9} & \underline{50.7}  & \underline{72.4} & \underline{50.1} & \underline{32.9} & \underline{67.4} \\
                          & MultiANDA               & \textbf{79.7}                    & \textbf{76.8}    & \textbf{64.8}     & \textbf{82.3}    & 49.3             & \textbf{34.4}    & \textbf{76.3}    \\
        \hline
        \multirow{10}{*}{\rotatebox[origin=c]{90}{VGG-19}}
                          & BIM                     & 9.6                              & 10.1             & 4.4               & 3.9              & 17.4             & 13.4             & 4.6              \\
                          & TIM                     & \underline{23.0}                 & \textbf{25.3}    & \textbf{17.5}     & 15.0             & \textbf{61.1}    & 18.0             & 20.1             \\
                          & SIM                     & 11.4                             & 12.0             & 5.6               & 8.7              & 18.4             & 13.4             & 7.4              \\
                          & DIM                     & 10.7                             & 10.7             & 4.7               & 5.6              & 17.9             & 13.6             & 5.9              \\
                          & FIA                     & 13.3                             & 12.2             & 8.3               & 7.1              & 23.7             & 14.6             & 9.8              \\
                          & TAIG                    & 17.7                             & 18.1             & 10.2              & \textbf{36.9}    & \underline{37.0} & \textbf{34.1}    & \textbf{41.9}    \\
                          & MI-FGSM                 & 13.0                             & 13.3             & 6.9               & 7.2              & 23.5             & 15.5             & 8.9              \\
                          & NI-FGSM                 & 13.9                             & 14.2             & 6.8               & 7.2              & 23.1             & 14.6             & 9.3              \\
                          & VMI-FGSM                & 21.9                             & 20.5             & 11.6              & 15.9             & 30.2             & 18.4             & 15.9             \\
                          & VNI-FGSM                & 21.6                             & 20.5             & 11.9              & 16.8             & 30.4             & \underline{18.6} & 15.7             \\\cline{2-9}
                          & ANDA                    & 20.9                             & 18.0             & 10.8              & 22.2             & 28.2             & 15.4             & 16.2             \\
                          & MultiANDA               & \textbf{24.6}                    & \underline{22.7} & \underline{13.7}  & \underline{31.2} & 27.9             & 16.2             & \underline{21.3} \\
        \hline
        \multirow{10}{*}{\rotatebox[origin=c]{90}{Inc-v3}}
                          & BIM                     & 11.1                             & 11.6             & 4.6               & 3.7              & 17.4             & 13.4             & 4.8              \\
                          & TIM                     & 27.5                             & 27.6             & 21.3              & 16.9             & \textbf{56.6}    & 22.8             & 21.1             \\
                          & SIM                     & 18.1                             & 18.6             & 8.4               & 8.6              & 20.1             & 15.2             & 10.8             \\
                          & DIM                     & 13.1                             & 13.3             & 6.7               & 5.8              & 18.2             & 12.8             & 8.6              \\
                          & FIA                     & 37.4                             & 36.7             & 21.3              & 11.6             & \underline{37.1} & 23.5             & 29.2             \\
                          & TAIG                    & 38.0                             & 36.8             & 23.9              & 22.8             & 31.5             & \textbf{29.6}    & 28.5             \\
                          & MI-FGSM                 & 18.3                             & 17.2             & 9.0               & 5.5              & 24.7             & 15.7             & 12.0             \\
                          & NI-FGSM                 & 18.6                             & 17.3             & 8.6               & 6.2              & 25.1             & 15.3             & 12.2             \\
                          & VMI-FGSM                & 36.9                             & 36.9             & 21.2              & 19.1             & 33.8             & 24.7             & 27.7             \\
                          & VNI-FGSM                & 36.4                             & 37.5             & 22.0              & 18.9             & 34.0             & \underline{25.3} & 27.4             \\\cline{2-9}
                          & ANDA                    & \underline{44.4}                 & \underline{43.0} & \underline{25.9}  & \underline{36.5} & 34.3             & 23.2             & \underline{37.0} \\
                          & MultiANDA               & \textbf{54.4}                    & \textbf{54.4}    & \textbf{36.7}     & \textbf{52.8}    & 32.3             & 24.3             & \textbf{46.9}    \\
        \bottomrule
    \end{tabular}
    \caption{\label{tab:defense_compl}
       Attack success rates (\%) of the proposed and baseline attacks against seven defense models. The best/second results are shown in bold/underlined.
    }
\end{table*}

\subsection{Cross-domain attack on CLIP}
\label{sec:clip}
Admittedly, using a surrogate with the identical dataset as target models for conducting black-box attacks is an ideal setting, typically only feasible in lab experiments or when targeting `general-purpose classification models' trained on ImageNet-based datasets, as in our case. 
We conducted an empirical verification of our methods on a `general-purpose classification model' trained on \textbf{non-ImageNet-based dataset}, CLIP~\cite{Rad21}.  Preliminary evaluation results, as shown in Table~\ref{tab:clip}, indicate that ANDA/MultiANDA also generalized well to such cross-domain model. 

\begin{table}[h]
    \small
    \centering
    \setlength\tabcolsep{5pt}
    \begin{tabular}{l|lll}
        \toprule
         \multirow{2}{*}{Attack} & \multicolumn{3}{c}{Source model}  \\
                                 & ResNet-50      & IncRes-v2        & VGG-19 \\ \hline                    
                           FIA   & 50.4           & 46.2             & 42.6 \\
                           TAIG  & 37.6           & 37.6             & 36.8 \\
                       VMI-FGSM  & 42.3	       & 45.4	          & 43.8 \\
                       \hline
                       ANDA      & \underline{51.6}       & \underline{49.6}	          & \underline{47.3}\\
                       MultiANDA & \textbf{53.0}	       & \textbf{50.6}	          & \textbf{48.5} \\                                             
        \bottomrule
    \end{tabular}
   \caption{\label{tab:clip}
        Success rates (\%)  on CLIP. Best results are shown in bold/underlined. The zero-shot performance of CLIP on clean ImageNet dataset is 73.7\%, i.e., the clean baseline (misclassification) is 26.3\%.
   }
\end{table}

%

\subsection{Transferability on ViT-based models}
Driven by the curiosity of whether the propose methods work well on  ViT-based architectures, we had conducted relating experiments using MultiANDA. Preliminary results in Table~\ref{vits} demonstrate notable transferability to ViTs~\cite{ViT}, also outperforming the selected baseline method.    

\subsection{Composite Transformation Attack}
In particular, we implemented the Composite Transformation (CT) Method with ANDA and MultiANDA (ANDA-CT and MultiANDA-CT). These input transformation techniques can synergize existing transfer-based attack methods, as demonstrated in SIM, DIM, and TIM. As shown by Wang \etal~\cite{Wang21a}, when integrated with VMI-FGSM and VNI-FGSM, VMI-CT-FGSM and VNI-CT-FGSM are recognized as state-of-the-art transferable-based attacks \cite{Wang21a}. Therefore, we focused our comparison to these methods. Our results, detailed in Table~\ref{tab:ct-normal} and Table~\ref{tab:ct-defense}, cover both normally trained and defense models, which demonstrate that ANDA-CT and MultiANDA-CT consistently enhance attack performance in almost all cases compared with the baselines. Specifically, 
our proposed methods have consistently achieved an average success rate increase of 3.5\% against black-box normally trained models and 3.6\% against advanced defense models.  This significant improvement highlights that the approximated posterior distribution over perturbations is more effective in crafting diverse adversarial examples than simply adopting data augmentation techniques.

\begin{table*}[h]
    \small
    \centering
    \begin{tabular}{ m{2cm} | m{2.2cm} m{2.2cm} m{2.2cm} m{2.2cm} }  
        \toprule
        \multirow{2}{*}{Source model } & \multicolumn{4}{c}{Target model}  \\                    
                  & ViT-L/16 \cite{ViT}            & DeiT3-B/16 \cite{DeiT}     & Swin-B/4 \cite{Swin}      & PiT-B             \cite{PiT} \\
        \hline
        ResNet-50 & 23.2/28.4/\textbf{30.5}  & 29.7/45.0/\textbf{46.6} & 24.9/38.7/\textbf{39.3} & 34.3/52.2/\textbf{56.3}  \\
        \hline
        IncRes-v2 & 29.0/28.4/\textbf{30.8}  & 30.9/41.5/\textbf{43.5} & 29.8/36.1/\textbf{38.3} & 38.6/49.9/\textbf{53.1} \\
        \hline
        VGG-19    & 14.7/\textbf{14.9}/14.8  & 16.5/16.7/\textbf{16.9} & 20.5/21.5/\textbf{22.1} & 24.9/26.7/\textbf{28.6}  \\
        \bottomrule
    \end{tabular}
    \caption{Success rates (\%) of VMI-FGSM/ANDA/MultiANDA on ViT-based target model. The best results are shown in bold.}           \label{vits}
\end{table*}

\begin{table*}[t]
    \renewcommand\arraystretch{1.25}
     \small
    \centering
    \begin{tabular}{c|c|rrrrrrr}
        \toprule
        \multirow{2}{*}{}                                    & \multirow{2}{*}{Attack} & \multicolumn{7}{c}{Target model}                                                                                                  \\\cline{3-9}
                                                             &                         & Inc-v3                           & Inc-v4        & ResNet-50     & ResNet-101    & ResNet-152    & IncRes-v2      & VGG-19        \\
        \hline
        \multirow{4}{*}{\rotatebox[origin=c]{90}{IncRes-v2}} & VNI-CT-FGSM             & 92.8                             & 91.7          & 88.7          & 89.0          & 87.8          & 99.5*          & 80.3          \\
                                                             & VMI-CT-FGSM             & 93.2                             & 91.5          & 88.6          & 88.7          & 87.5          & 99.2*          & 81.8          \\\cline{2-9}
                                                             & ANDA-CT                 & 96.7                             & 95.8          & 94.5          & \textbf{94.5} & 93.9          & \textbf{99.8*} & \textbf{87.9} \\
                                                             & MultiANDA-CT            & \textbf{96.8}                    & \textbf{96.3} & \textbf{94.6} & 94.4          & \textbf{94.3} & 99.7*          & 87.5          \\
        \hline
        \multirow{4}{*}{\rotatebox[origin=c]{90}{ResNet-50}} & VNI-CT-FGSM             & 91.0                             & 88.6          & \textbf{100*} & 97.7          & 97.6          & 87.7           & 79.5          \\
                                                             & VMI-CT-FGSM             & 91.1                             & 88.4          & \textbf{100*} & 97.6          & 97.2          & 88.4           & 80.0          \\\cline{2-9}
                                                             & ANDA-CT                 & 95.6                             & \textbf{94.5} & \textbf{100*} & 99.1          & \textbf{98.9} & 94.4           & \textbf{87.0} \\
                                                             & MultiANDA-CT            & \textbf{96.1}                    & \textbf{94.5} & \textbf{100*} & \textbf{99.2} & 98.7          & \textbf{95.4}  & 85.5          \\
        \hline
        \multirow{4}{*}{\rotatebox[origin=c]{90}{VGG-19}}    & VNI-CT-FGSM             & 84.7                             & 87.9          & 81.7          & 75.2          & 72.4          & 77.2           & \textbf{100*} \\
                                                             & VMI-CT-FGSM             & 84.9                             & 88.9          & 81.9          & \textbf{75.3} & 72.9          & 77.4           & \textbf{100*} \\\cline{2-9}
                                                             & ANDA-CT                 & 83.7                             & 89.4          & 81.4          & 74.3          & 73.0          & 77.5           & \textbf{100*} \\
                                                             & MultiANDA-CT            & \textbf{85.5}                    & \textbf{89.6} & \textbf{82.3} & 74.7          & \textbf{73.4} & \textbf{78.1}  & \textbf{100*} \\
        \hline
        \multirow{4}{*}{\rotatebox[origin=c]{90}{Inc-v3}}    & VNI-CT-FGSM             & 99.9*                            & 91.2          & 85.8          & 83.8          & 82.8          & 88.2           & 81.3          \\
                                                             & VMI-CT-FGSM             & 99.8*                            & 91.5          & 86.3          & 83.6          & 82.3          & 88.2           & 80.5          \\\cline{2-9}
                                                             & ANDA-CT                 & \textbf{100*}                    & 95.0          & 90.1          & \textbf{88.3} & 87.6          & 93.6           & \textbf{84.9} \\
                                                             & MultiANDA-CT            & \textbf{100*}                    & \textbf{95.7} & \textbf{90.4} & 88.1          & \textbf{88.9} & \textbf{93.7}  & 82.8          \\

        \bottomrule
    \end{tabular}
    \caption{\label{tab:ct-normal}
        Attack success rates (\%) against seven normally trained models using the proposed method and the various selected attacks enhanced by CT with corresponding source models. The best results are shown in bold. Sign * indicates the results against white-box models.
    }
\end{table*}

\begin{table*}[t]
    \renewcommand\arraystretch{1.35}
     \small
    \centering
    \begin{tabular}{c|c|ccccccc}
        \toprule
        \multirow{2}{*}{}                                    & \multirow{2}{*}{Attack} & \multicolumn{7}{c}{Target model}                                                                                                       \\\cline{3-9}
                                                             &                         & Inc-v3$_{ens3}$                  & Inc-v3$_{ens4}$ & IncRes-v2$_{ens}$ & HGD           & RS            & NRP           & NIPS-r3       \\
        \hline
        \multirow{4}{*}{\rotatebox[origin=c]{90}{IncRes-v2}} & VNI-CT-FGSM             & 86.9                             & 85.8            & 83.4              & 84.3          & 71.1          & 73.3          & 84.8          \\
                                                             & VMI-CT-FGSM             & 86.7                             & 85.1            & 83.8              & 83.8          & 71.3          & 74.4          & 84.3          \\\cline{2-9}
                                                             & ANDA-CT                 & 92.5                             & 91.1            & 89.1              & 89.3          & \textbf{77.9} & 81.1          & 91.3          \\
                                                             & MultiANDA-CT            & \textbf{93.6}                    & \textbf{93.1}   & \textbf{91.7}     & \textbf{92.1} & 76.9          & \textbf{81.5} & \textbf{92.9} \\
        \hline
        \multirow{4}{*}{\rotatebox[origin=c]{90}{ResNet-50}} & VNI-CT-FGSM             & 84.7                             & 82.6            & 77.6              & 81.4          & 76.2          & 73.4          & 82.0          \\
                                                             & VMI-CT-FGSM             & 86.1                             & 83.3            & 76.0              & 82.1          & 76.8          & 72.6          & 81.9          \\\cline{2-9}
                                                             & ANDA-CT                 & 91.9                             & 88.0            & 81.9              & 89.3          & 80.2          & \textbf{78.8} & 88.5          \\
                                                             & MultiANDA-CT            & \textbf{92.5}                    & \textbf{90.4}   & \textbf{84.7}     & \textbf{90.1} & \textbf{81.0} & 78.1          & \textbf{89.7} \\
        \hline
        \multirow{4}{*}{\rotatebox[origin=c]{90}{VGG-19}}    & VNI-CT-FGSM             & 55.3                             & \textbf{54.8}   & 40.6              & 54.8          & \textbf{65.9} & 37.5          & 51.2          \\
                                                             & VMI-CT-FGSM             & 55.8                             & 54.0            & 41.1              & 54.9          & 65.7          & \textbf{37.6} & 51.5          \\\cline{2-9}
                                                             & ANDA-CT                 & 50.2                             & 52.4            & 39.9              & 53.6          & 62.1          & 33.0          & 48.1          \\
                                                             & MultiANDA-CT            & \textbf{57.0}                    & 54.2            & \textbf{42.1}     & \textbf{58.8} & 62.2          & 33.9          & \textbf{53.8} \\
        \hline
        \multirow{4}{*}{\rotatebox[origin=c]{90}{Inc-v3}}    & VNI-CT-FGSM             & 81.0                             & 80.5            & 68.4              & 72.4          & 64.9          & 63.4          & 73.6          \\
                                                             & VMI-CT-FGSM             & 81.7                             & 79.0            & 66.2              & 72.1          & 64.9          & \textbf{65.0} & 73.2          \\\cline{2-9}
                                                             & ANDA-CT                 & 84.9                             & 81.6            & 67.8              & 77.0          & 66.5          & 62.6          & 78.2          \\
                                                             & MultiANDA-CT            & \textbf{88.4}                    & \textbf{84.9}   & \textbf{74.7}     & \textbf{82.3} & \textbf{66.9} & 64.5          & \textbf{81.6} \\

        \bottomrule
    \end{tabular}
    \caption{\label{tab:ct-defense}
        Attack success rates (\%) against seven advanced defense models using the proposed method and the various selected attacks enhanced by CT techniques with corresponding source models. The best results are shown in bold.
    }
\end{table*}

\subsection{Comprehensive Comparison with TAIG}
\label{sec:taig}
In the experiments illustrated in Table~\ref{tab:normaltrained_compl} and Table~\ref{tab:defense_compl}, we used the same parameter settings for all baselines including TAIG and our proposed methods for a fair comparison. To showcase TAIG's best performance, we replicated the experiments following TAIG's official settings. In particular, we set $\epsilon=0.03, 0.05, 0.1$ and $T=20, 50, 100$. The results presented in Table~\ref{tab:taig_normal} shows the exceptional transferability performance of adversarial examples generated by ANDA across all settings. Although there are some variations in effectiveness against advanced defense models as shown in Table~\ref{tab:taig_defense}, our proposed method secures the best results for every target model in almost all tested scenarios.

\begin{table*}[h]
    \small
    \centering
    \begin{tabular}{c|c|rrrrrrr}
        \toprule
        \multirow{1}{*}{Settings} & \diagbox{Source}{Target} & Inc-v3 & Inc-v4                  & ResNet-50                        & ResNet-101           & ResNet-152           & IncRes-v2            & VGG-19                           \\
        \midrule
        \multirow{4}{*}{\rotatebox[origin=c]{90}{\thead{$\epsilon=0.03$\\ $T=20$}}} 
        & IncRes-v2                & \textbf{82.0}/49.5               & \textbf{79.2} /41.2     & \textbf{70.4} /39.9             & \textbf{66.4} /37.9   & \textbf{65.5} /35.5  & \textbf{99.2} /93.0* & \textbf{66.5} /38.6             \\
        & ResNet-50                & \textbf{82.8}  /37.0             & \textbf{76.8} /29.0     & 99.7 /\textbf{100.0*}            & \textbf{95.0} /67.3   & \textbf{93.9} /62.3  & \textbf{76.8} /23.8 & \textbf{73.6} /40.3             \\
        & VGG-19                   & \textbf{46.8}  /29.3             & \textbf{53.6} /27.4     & \textbf{39.6} /26.4             & \textbf{35.1} /22.5   & \textbf{31.7} /18.2  & \textbf{31.6} /15.5 & \textbf{100.0} /\textbf{100.0*} \\
        & Inc-v3                   & \textbf{99.9}  /99.3*             & \textbf{67.4} /34.9     & \textbf{54.6} /30.5             & \textbf{50.7} /28.6   & \textbf{46.7} /24.0  & \textbf{58.0} /27.6 & \textbf{58.6} /36.7             \\
\hline    
\multirow{4}{*}{\rotatebox[origin=c]{90}{\thead{$\epsilon=0.05$\\ $T=50$}}} 
        & IncRes-v2                & \textbf{90.4} /66.8             & \textbf{88.7} /60.0 & \textbf{82.7} /55.6             & \textbf{80.8} /52.5 & \textbf{79.2} /49.8 & \textbf{99.8} /95.5* & \textbf{78.2} /48.4             \\
        & ResNet-50                & \textbf{78.5}  /54.5             & \textbf{75.6} /46.1 & 92.8 /\textbf{100.0*}            & \textbf{87.1} /85.1 & \textbf{85.8} /81.6 & \textbf{74.3} /42.4 & \textbf{73.9} /50.9             \\
        & VGG-19                   & \textbf{68.0}  /42.7             & \textbf{73.7} /44.7 & \textbf{57.9} /37.6             & \textbf{52.4} /33.2 & \textbf{48.7} /28.7 & \textbf{54.2} /27.9 & \textbf{100.0}  /\textbf{100.0*} \\
        & Inc-v3                   & \textbf{100.0} /99.7*            & \textbf{77.6} /51.1 & \textbf{68.3} /45.7             & \textbf{65.0} /41.7 & \textbf{61.4} /36.2 & \textbf{72.4} /45.8 & \textbf{69.0} /45.3             \\
\hline
\multirow{4}{*}{\rotatebox[origin=c]{90}{\thead{$\epsilon=0.1$\\ $T=100$}}} 
        & IncRes-v2                & \textbf{95.9} /80.7             & \textbf{95.0} /75.6 & \textbf{90.7} /70.7             & \textbf{90.2} /69.3 & \textbf{91.1} /66.3 & \textbf{99.9} /97.6* & \textbf{89.3} /61.8             \\
        & ResNet-50                & \textbf{98.2} /71.6             & \textbf{97.2} /62.0 & \textbf{100.0}  /\textbf{100.0*} & \textbf{99.9} /94.4 & \textbf{99.8} /91.8 & \textbf{97.3} /63.2 & \textbf{94.1} /64.6             \\
        & VGG-19                   & \textbf{87.7} /59.3             & \textbf{92.2} /63.7 & \textbf{81.5} /55.9             & \textbf{78.6} /45.7 & \textbf{77.0} /44.6 & \textbf{80.5} /44.0 & \textbf{100.0}  /\textbf{100.0*} \\
        & Inc-v3                   & \textbf{100.0}  /\textbf{100.0*} & \textbf{88.7} /67.6 & \textbf{83.0} /59.1             & \textbf{80.0} /53.7 & \textbf{78.8} /53.4 & \textbf{86.7} /64.1 & \textbf{82.9} /55.9             \\
        \bottomrule
    \end{tabular}
    \caption{\label{tab:taig_normal}
        Attack success rates (\%) against seven normally trained models using the proposed method and TAIG, under official TAIG settings (denoted as ANDA/TAIG). The higher success rate of each comparison is highlighted in bold. Sign * indicates the results on white-box models.
    }
\end{table*}

\begin{table*}[h]
    \small
    \centering
    \begin{tabular}{c|c|rrrrrrr}
        \toprule
        \multirow{1}{*}{Settings} & \diagbox{Source}{Target} & Inc-v3$_{ens3}$      & Inc-v3$_{ens4}$      & IncRes-v2$_{ens}$    & HGD                  & RS                   & NRP                  & NIPS-r3              \\
        \midrule

        \multirow{4}{*}{\rotatebox[origin=c]{90}{\thead{$\epsilon=0.03$\\ $T=20$}}} 
                                     & IncRes-v2                & \textbf{45.5} /32.3 & \textbf{42.8} /31.8 & \underline{\textbf{34.1}} /25.3 & \textbf{42.7} /21.6 & \textbf{25.1} /24.8 & 18.3 /\textbf{20.2} & \textbf{39.7} /23.4 \\
                                     & ResNet-50                & \underline{\textbf{47.2}} /23.3 & \underline{\textbf{47.3}} /25.2 & \textbf{31.8} /15.8 & \underline{\textbf{50.0}} /14.7 & \underline{\textbf{28.4}} /25.0 & \underline{\textbf{20.9}} /18.3 & \underline{\textbf{41.8}} /16.7 \\
                                     & VGG-19                   & \textbf{11.4} /11.3 & 13.1 /\textbf{14.3} & 6.4 /\textbf{6.7}   & \textbf{11.1} /7.5  & 19.2 /\textbf{20.8} & 12.8 /\textbf{13.8} & \textbf{8.5} /7.4   \\
                                     & Inc-v3                   & \textbf{27.3} /24.7 & \textbf{29.6} /26.8 & \textbf{14.0} /12.8 & \textbf{21.3} /12.3 & 21.5 /\textbf{22.6} & 15.8 /\textbf{19.2} & \textbf{18.4} /13.5 \\
        \hline
        \multirow{4}{*}{\rotatebox[origin=c]{90}{\thead{$\epsilon=0.05$\\ $T=50$}}} 
                                    & IncRes-v2                 & \underline{\textbf{58.1}} /48.6 & \underline{\textbf{55.1}} /47.6 & \underline{\textbf{43.1}} /42.1 & \underline{\textbf{55.0}} /36.2 & \underline{\textbf{35.1}} /33.0 & 24.2 /\underline{\textbf{29.4}} & \underline{\textbf{53.8}} /37.4 \\
                                     & ResNet-50                & \textbf{48.7} /38.8 & \textbf{46.7} /37.8 & \textbf{32.2} /28.3 & \textbf{48.1} /28.4 & \underline{\textbf{35.1}} /32.4 & 22.6 /\textbf{25.3} & \textbf{44.7} /28.1 \\
                                     & VGG-19                   & 14.6 /\textbf{18.3} & 15.0 /\textbf{18.4} & 9.0 /\textbf{11.6}  & \textbf{18.4} /13.1 & 22.2 /\textbf{24.0} & 13.5 /\textbf{14.8} & 12.4 /\textbf{12.9} \\
                                     & Inc-v3                   & 36.2 /\textbf{36.5} & 35.0 /\textbf{36.0} & 17.9 /\textbf{22.9} & \textbf{26.8} /22.3 & 27.0 /\textbf{28.0} & 19.3 /\textbf{25.9} & \textbf{27.2} /25.5 \\
        \hline
        \multirow{4}{*}{\rotatebox[origin=c]{90}{\thead{$\epsilon=0.1$\\ $T=100$}}} 
                                    & IncRes-v2                 & \textbf{72.3} /68.6 & \textbf{68.0} /65.7 & 59.1 /\textbf{61.2} & \textbf{69.5} /56.8 & \textbf{66.1} /50.1 & 42.1 /\textbf{49.1} & \underline{\textbf{69.4}} /58.1 \\
                                     & ResNet-50                & \underline{\textbf{81.6}} /59.1 & \underline{\textbf{76.4}} /58.1 & \underline{\textbf{62.7}} /46.7 & \underline{\textbf{83.6}} /49.5 & \underline{\textbf{71.8}} /52.9 & \underline{\textbf{78.0}} /46.4 & 43.2 /\textbf{48.8} \\
                                     & VGG-19                   & 28.5 /\textbf{29.2} & 26.3 /\textbf{29.7} & 17.5 /\textbf{19.1} & \textbf{38.5} /24.8 & \textbf{45.4} /32.7 & \textbf{20.2} /19.4 & \textbf{28.6} /22.9 \\
                                     & Inc-v3                   & 45.3 /\textbf{50.9} & 45.4 /\textbf{52.2} & 28.6 /\textbf{35.4} & \textbf{36.4} /34.3 & \textbf{47.8} /39.3 & 28.8 /\textbf{39.3} & \textbf{41.3} /38.6 \\
        \bottomrule
    \end{tabular}
    \caption{\label{tab:taig_defense}
     Attack success rates (\%) against seven defense models using the proposed method and TAIG, under official TAIG settings (denoted as ANDA/TAIG). The higher success rate of each comparison is highlighted in bold. The best result for every target model within each setting is underlined.
    }
\end{table*}

\subsection{Perturbation Visualization}
\begin{figure*}[ht]
    \centering
    \begin{minipage}[t]{0.48\linewidth}
        \centering
        \includegraphics[width=0.95\linewidth]{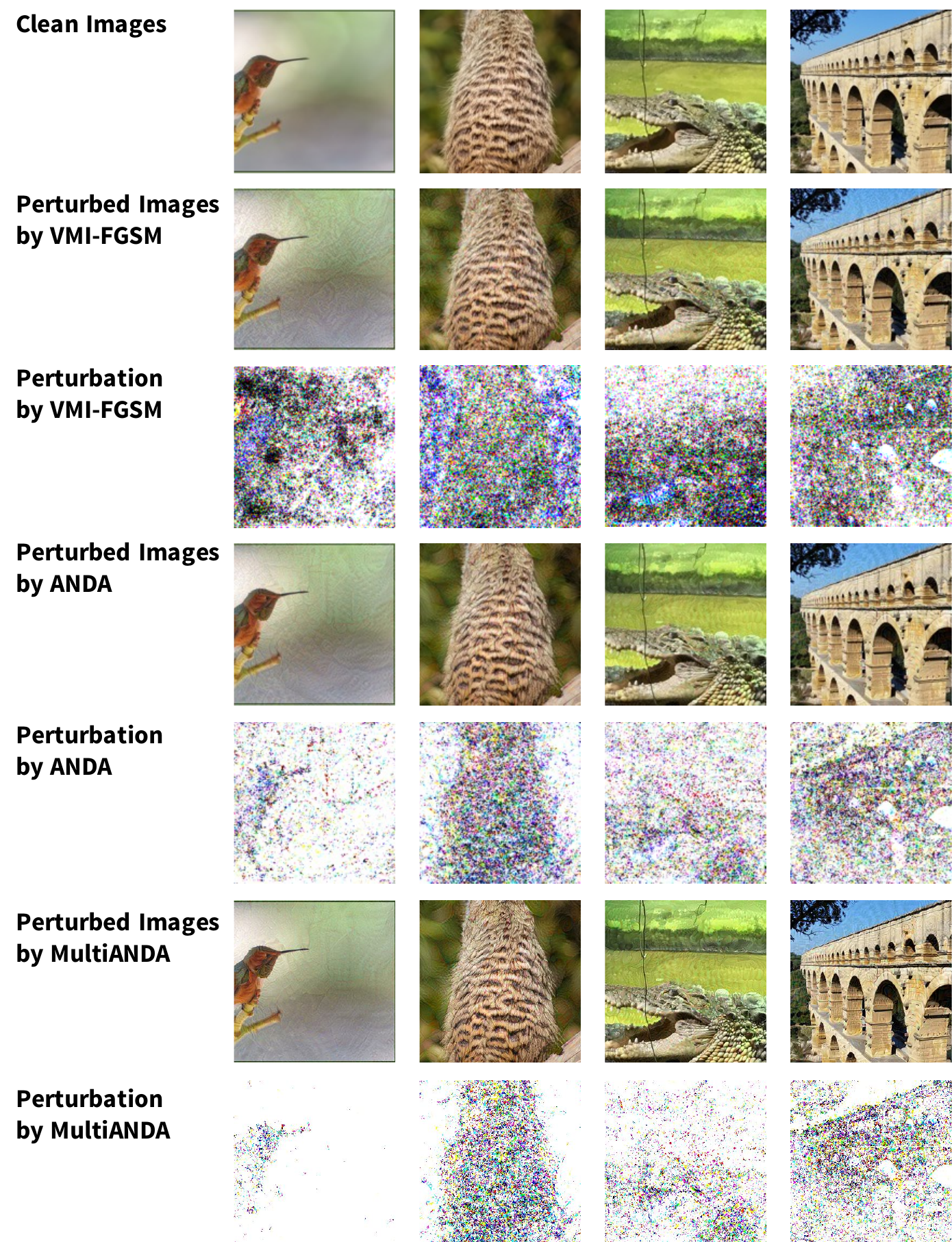}
        \caption{Perturbation visualization of the examples that fool the normally trained models}
        \label{fig:merge_normal}
    \end{minipage} \hfill
    \begin{minipage}[t]{0.48\linewidth}
        \centering
        \includegraphics[width=1\linewidth]{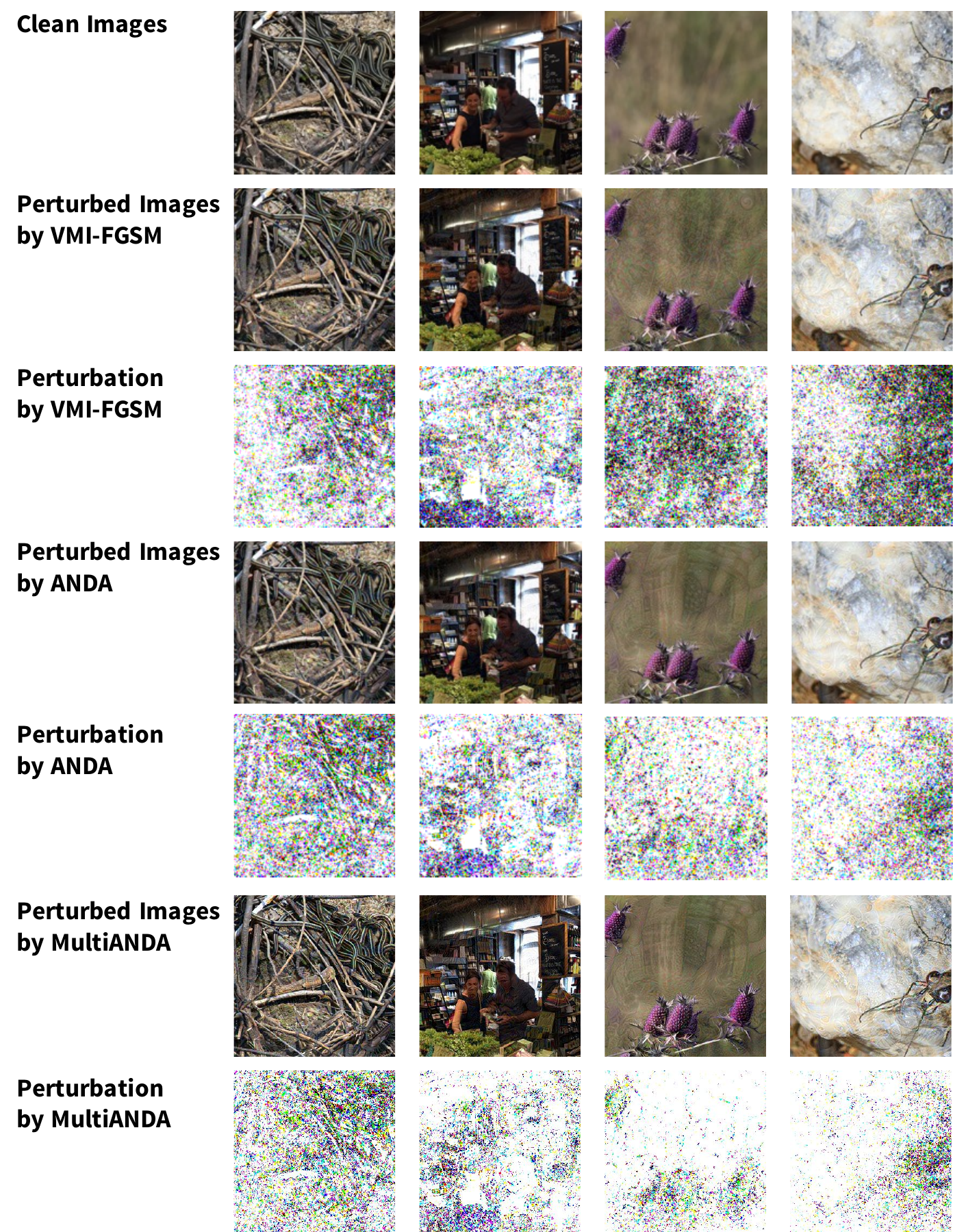}
        \caption{Perturbation visualization of the examples that fool the defense models}
        \label{fig:merge_defense}
    \end{minipage}
\end{figure*}

In previous sections, we compared the attack success rates of our methods with baseline approaches. This section offers a visual analysis of the adversarial perturbations created to assess their effects. We select and display images where adversarial examples generated by ANDA and MultiANDA successfully attack all target models, , in contrast to those by VMI-FGSM, which failed to deceive any. These experiments were conducted across both normally trained black-box models and various defense models. Figures~\ref{fig:merge_normal} and \ref{fig:merge_defense} demonstrate examples that successfully fooled six black-box models and seven defense models, respectively.

The illustrations reveal that the perturbations generated by ANDA and MultiANDA (Rows 5 and 7) target the semantic features of objects more precisely than VMI-FGSM (Row 3). For instance, in the first column of Figure~\ref{fig:merge_normal} and the third column of Figure~\ref{fig:merge_defense}, ANDA and MultiANDA  impose perturbations to decisive regions, such as the bird's head and the flowers, which are crucial for model prediction. This indicates that our methods precisely approximate the optimal solution for the maximization problem introduced in Section 2.1, as alterations in these key areas significantly increase the loss values. Furthermore, the pronounced denoising effect shown in the visualizations of perturbations by MultiANDA highlights the enhanced performance enabled by the mixture of Gaussian models.

\subsection{Time and Memory Analysis}

After thorough analysis of the proposed algorithm, we found that the computational overhead largely depends on the number of batch samples during augmentations. The cost of collecting statistical information within our method was negligible.
We conducted experimental comparisons with the baseline methods that also employed augmentations, under the same settings as in Section~\ref{sec:hp}. As shown in the Table~\ref{tab:time_tab} (mean values of five repeat trials), ANDA even consumes less computational time than the state-of-the-art baselines, VMI-FGSM and VNI-FGSM.
Although ANDA incurs additional memory overhead for the covariance matrix, the total memory usage is still approximately 1.6\% lower than that of FIA.

MultiANDA's implementation involves repeating the ANDA process multiple times; hence, the time cost is roughly $K$ times that of a single ANDA, where $K$ is the number of ensembled ANDAs. However, as the ANDAs in MultiANDA operate independently, distributed computing technology can significantly accelerate the process. For ANDA, all experiments were conducted on a single GPU (NVIDIA Geforce RTX 2080 Ti), while for MultiANDA, we utilized $K$ GPUs to enable distributed processing.
Furthermore, when the number of samples per batch augmentation is relatively small ($n \leq 36$), batch processing can be applied to expedite our algorithms. For detailed implementation, please refer to our publicly released code.

\begin{table}[ht]
    \centering
    \includegraphics[width=0.95\linewidth]{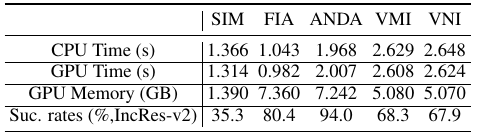}
    \caption{CPU Time, GPU Time, GPU Memory(GB) and attacking performance of different methods. The source model and target model are ResNet-50 and IncRes-v2, respectively.}
    \label{tab:time_tab}
\end{table}

\subsection{Optimization Trajectory}

\begin{figure*}[ht]
    \centering
    \begin{minipage}[t]{0.48\linewidth}
        \centering
        \begin{subfigure}[t]{1\textwidth}
            \centering
            \includegraphics[width=\textwidth]{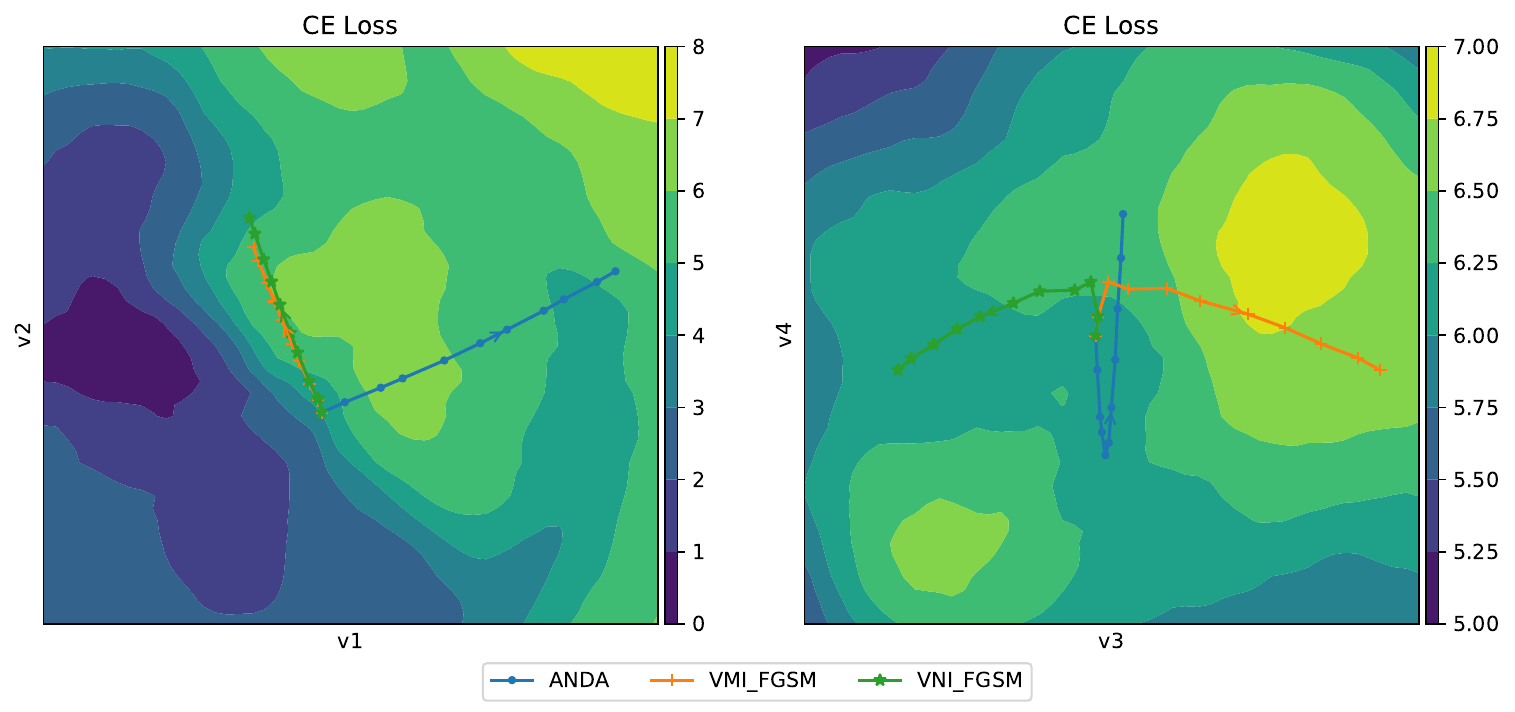}
            \caption{Inc-v4}
        \end{subfigure}
        \hfill
        \begin{subfigure}[t]{1\textwidth}
            \centering
            \includegraphics[width=\textwidth]{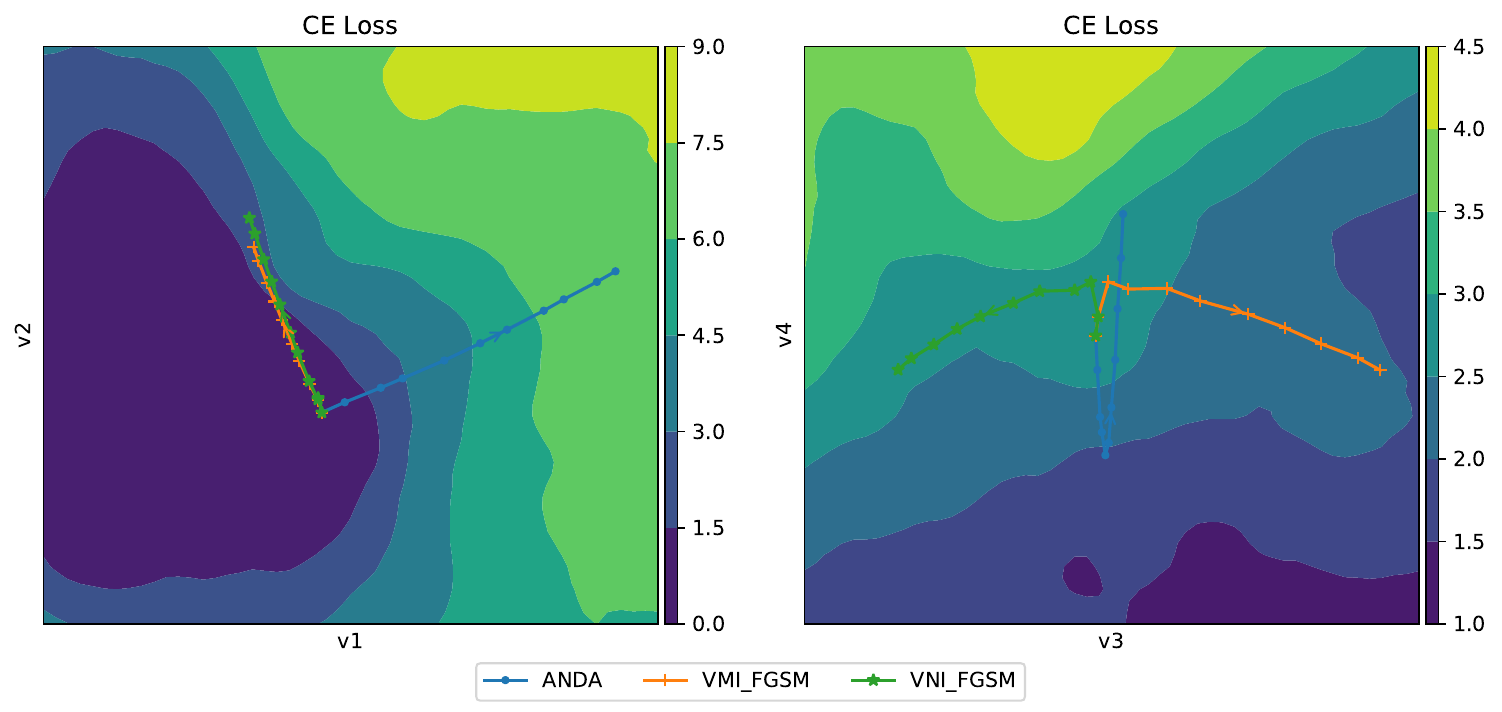}
            \caption{IncRes-v2}
        \end{subfigure}

        \begin{subfigure}[t]{1\textwidth}
            \centering
            \includegraphics[width=\textwidth]{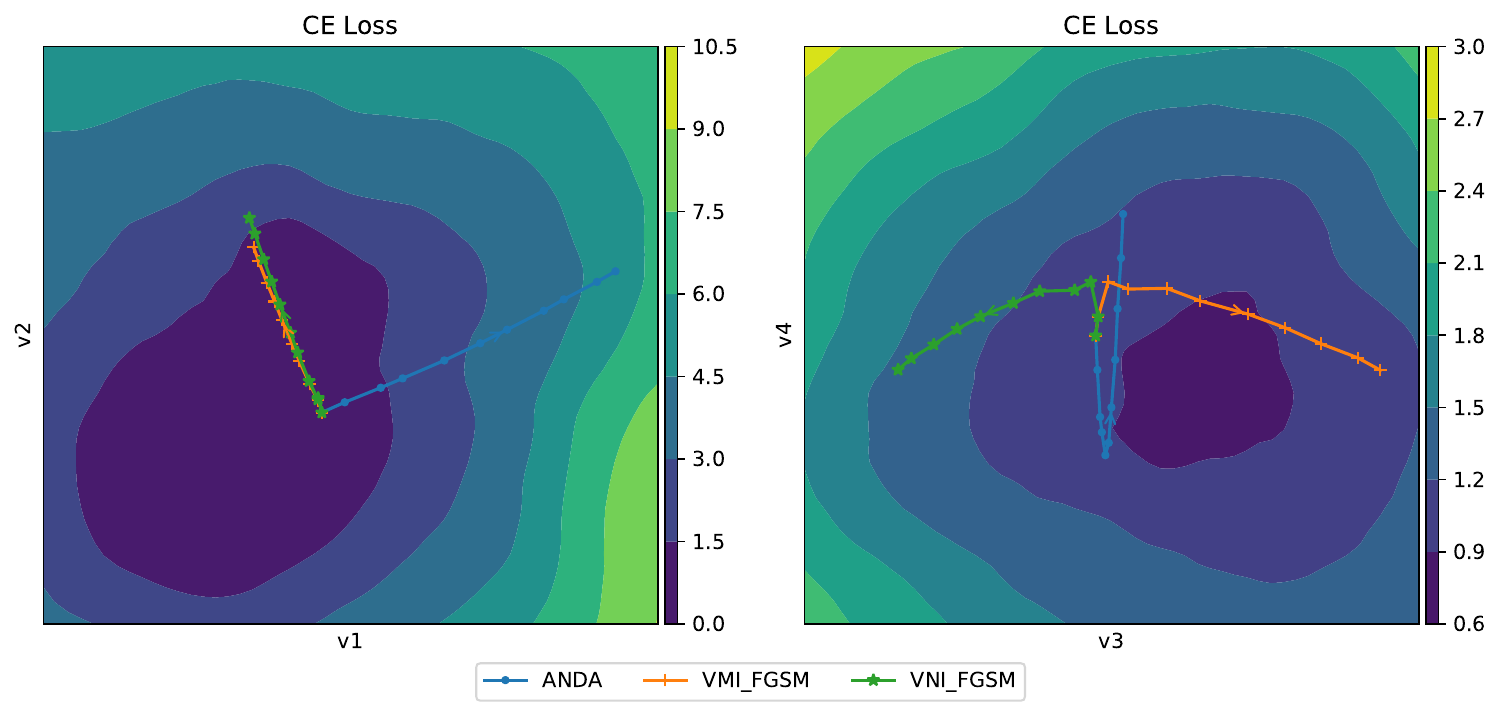}
            \caption{VGG-19}
        \end{subfigure}
        \hfill
        \begin{subfigure}[t]{1\textwidth}
            \centering
            \includegraphics[width=\textwidth]{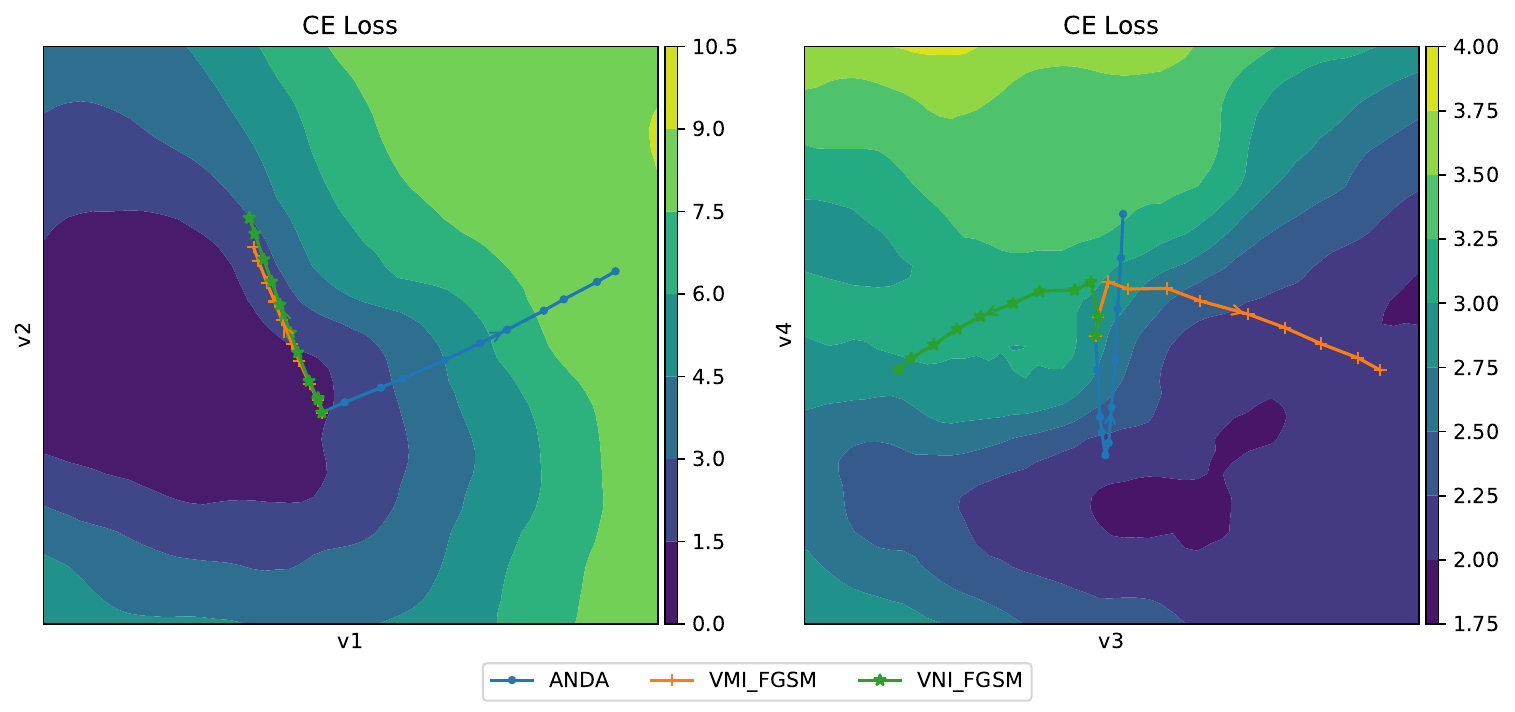}
            \caption{ResNet-50}
        \end{subfigure}

    \end{minipage} \hfill
    \begin{minipage}[t]{0.48\linewidth}
        \centering
        \begin{subfigure}[t]{1\textwidth}
            \centering
            \includegraphics[width=\textwidth]{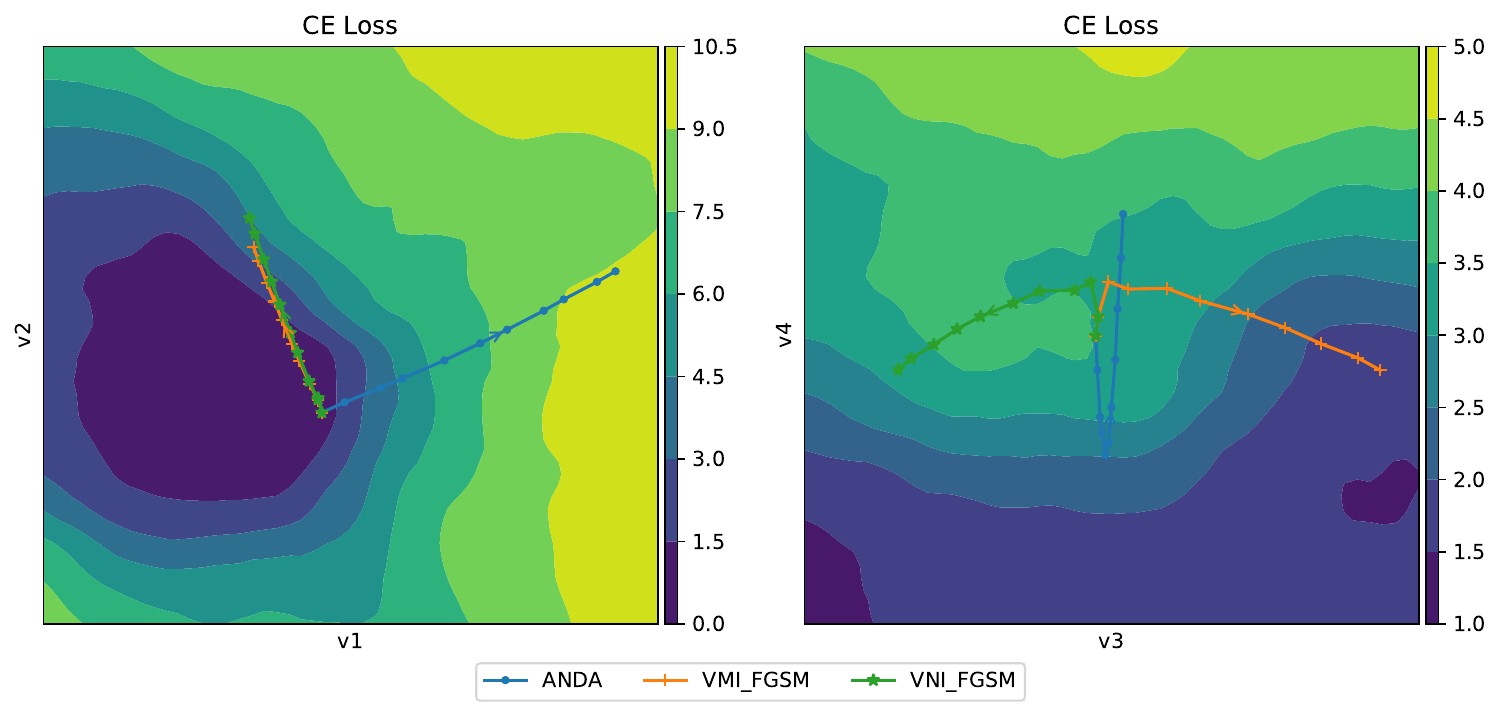}
            \caption{ResNet-101}
        \end{subfigure}
        \hfill
        \begin{subfigure}[t]{1\textwidth}
            \centering
            \includegraphics[width=\textwidth]{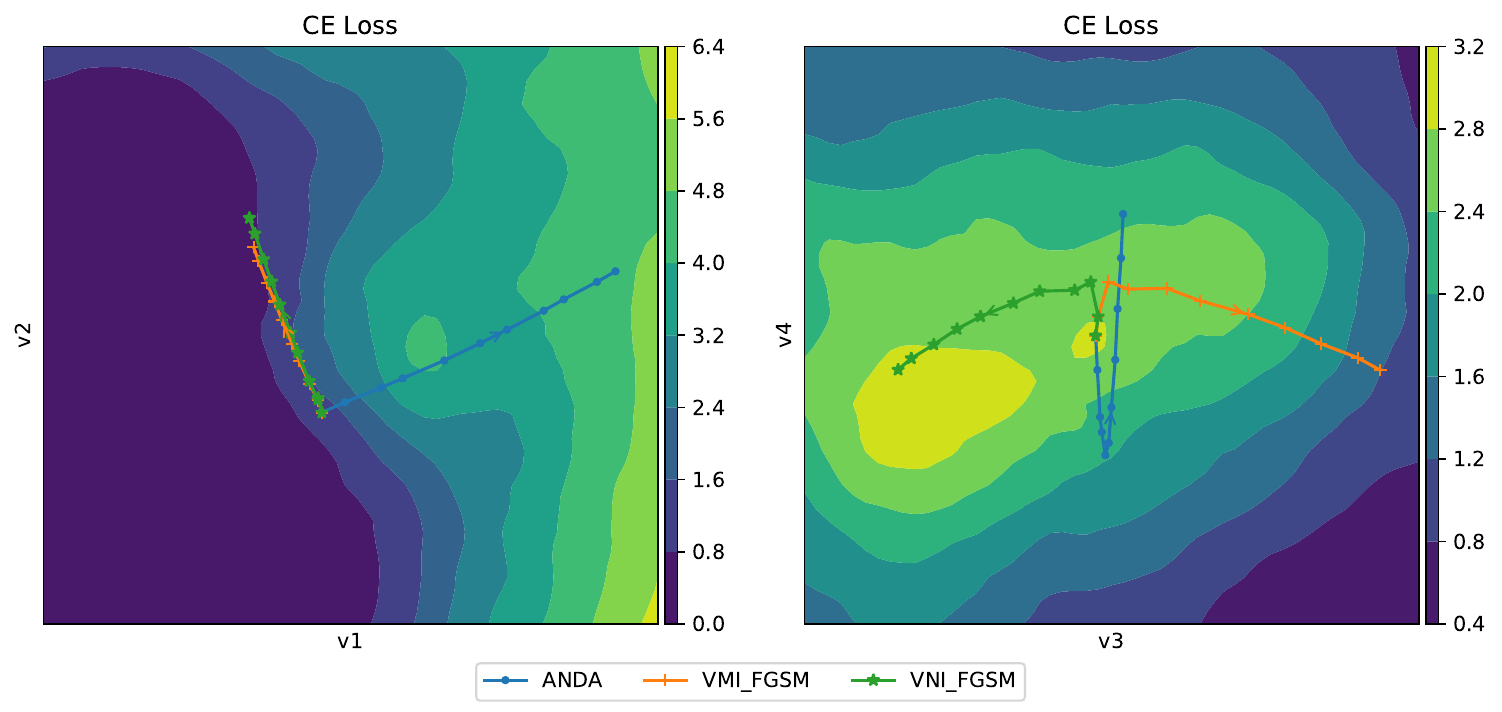}
            \caption{IncRes-v2$_{ens}$}
        \end{subfigure}

        \begin{subfigure}[t]{1\textwidth}
            \centering
            \includegraphics[width=\textwidth]{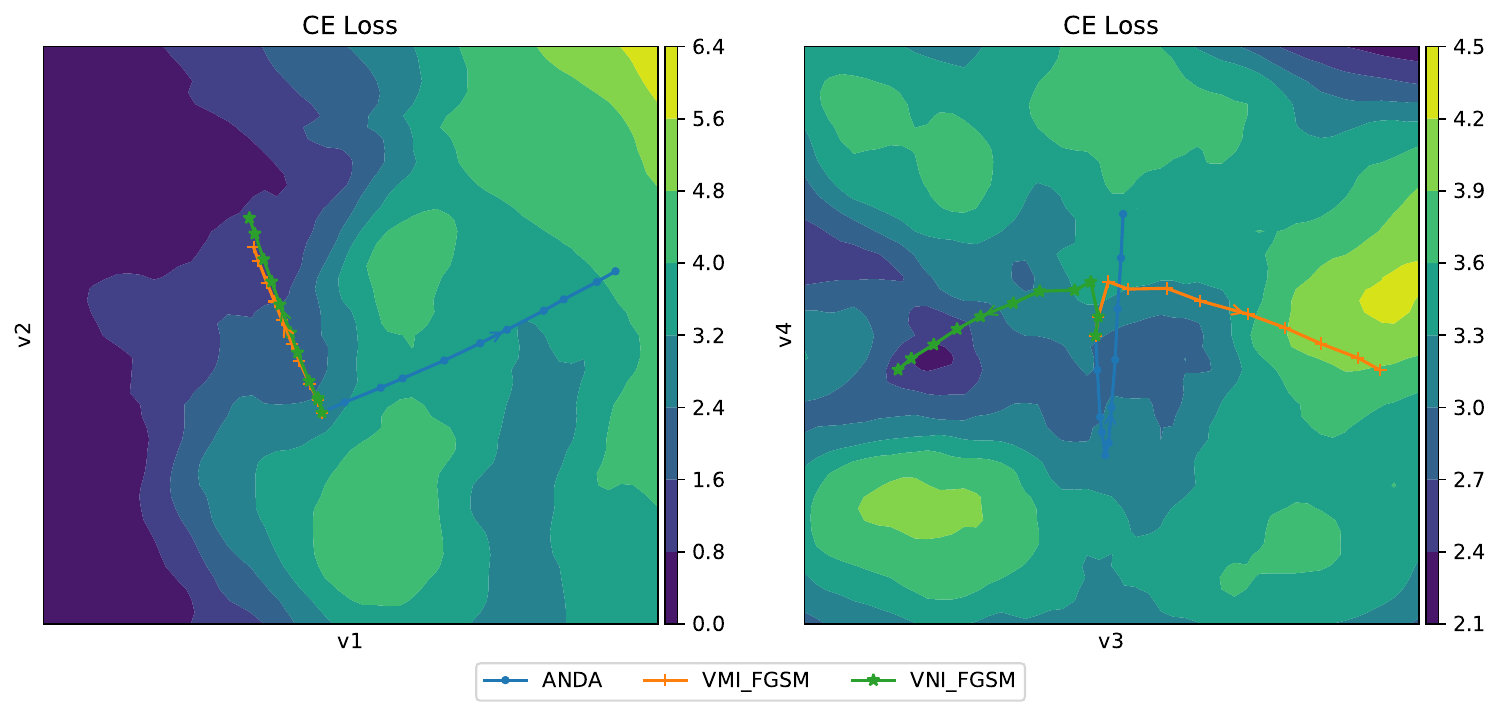}
            \caption{Inc-v3$_{ens3}$}
        \end{subfigure}
        \hfill
        \begin{subfigure}[t]{1\textwidth}
            \centering
            \includegraphics[width=\textwidth]{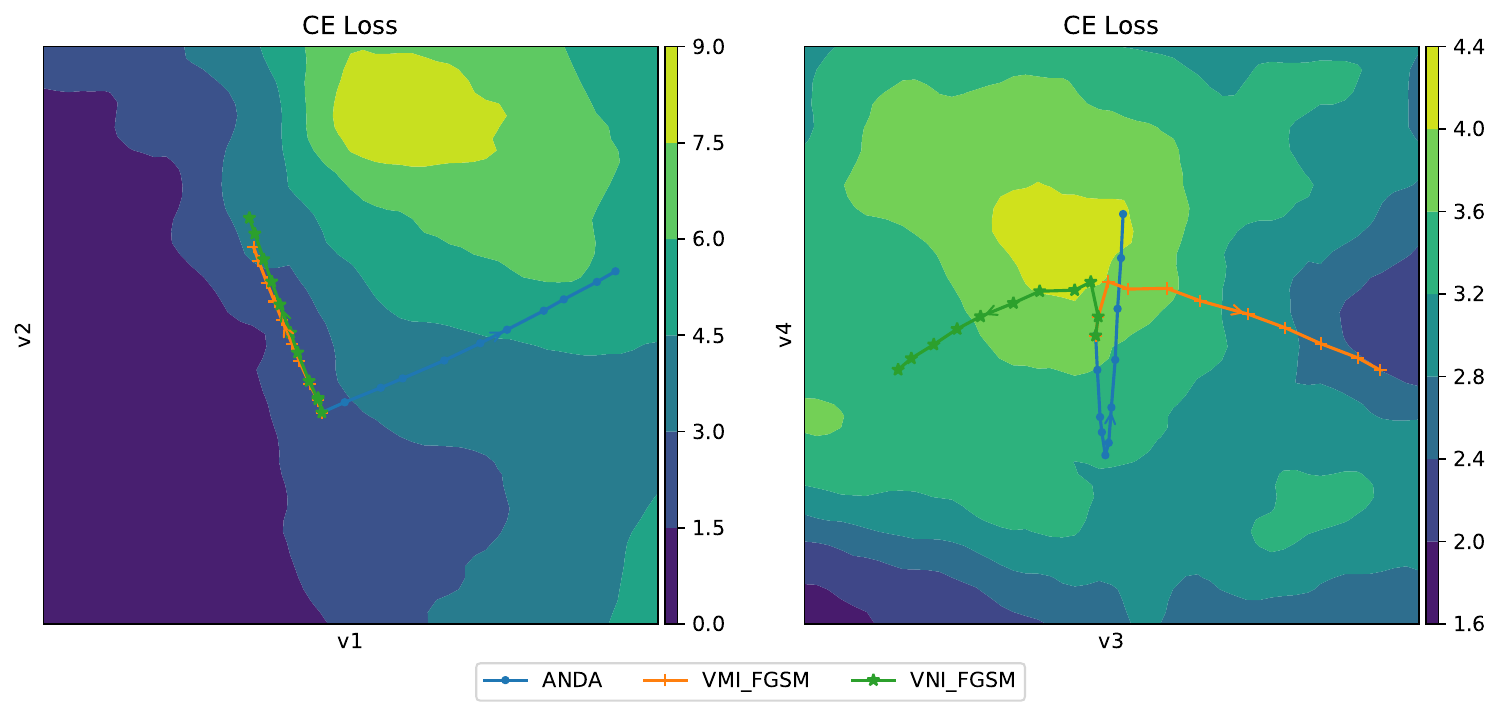}
            \caption{Inc-v3$_{ens4}$}
        \end{subfigure}
    \end{minipage}
    \caption{Visualization of optimization trajectories for generated adversarial examples by ANDA, VMI-FGSM, and VNI-FGSM}
    \label{fig:opt_trajectory}
\end{figure*}

To evaluate the optimization efficacy of ANDA, we analyzed the optimization trajectories of ANDA, and baseline methods VMI-FGSM and VNI-FGSM. For enhanced visualization, we employed Principal Component Analysis (PCA)~\cite{li2018visualizing} and projected the iteratively optimized samples (including original images and their corresponding adversarial examples of 10 iterations against Inc-v3) along the top four principal components (v1-v2 and v3-v4). We randomly chose the initial inputs and selected the ones that succeeded in deceiving the target black-box models. We visualized the iteration paths of these algorithms on the 2D loss landscape against eight black-box models. The results, as depicted in Figure~\ref{fig:opt_trajectory}, demonstrate that ANDA finds more efficient optimization paths than VMI/VNI-FGSM, leading to enhanced optimization outcomes.

\section{Ablation Study}
\label{appen:ablat}
In ANDA, we integrate two components, namely data augmentations and the approximation of perturbation distribution, to enhance the transferability of adversarial examples. This section examines the individual contributions of these components to ANDA's attack performance and the impact of key hyper-parameters. Note that white-box attack performance may reach 100\%. Therefore, we focus on black-box settings to precisely evaluate the efficacy of each component. We select seven representative models, including four normally and three adversarially trained models, as targets.

\subsection{Number of Augmentation Batches}
\begin{figure*}[ht]
    \centering
    \includegraphics[width=0.9\linewidth]{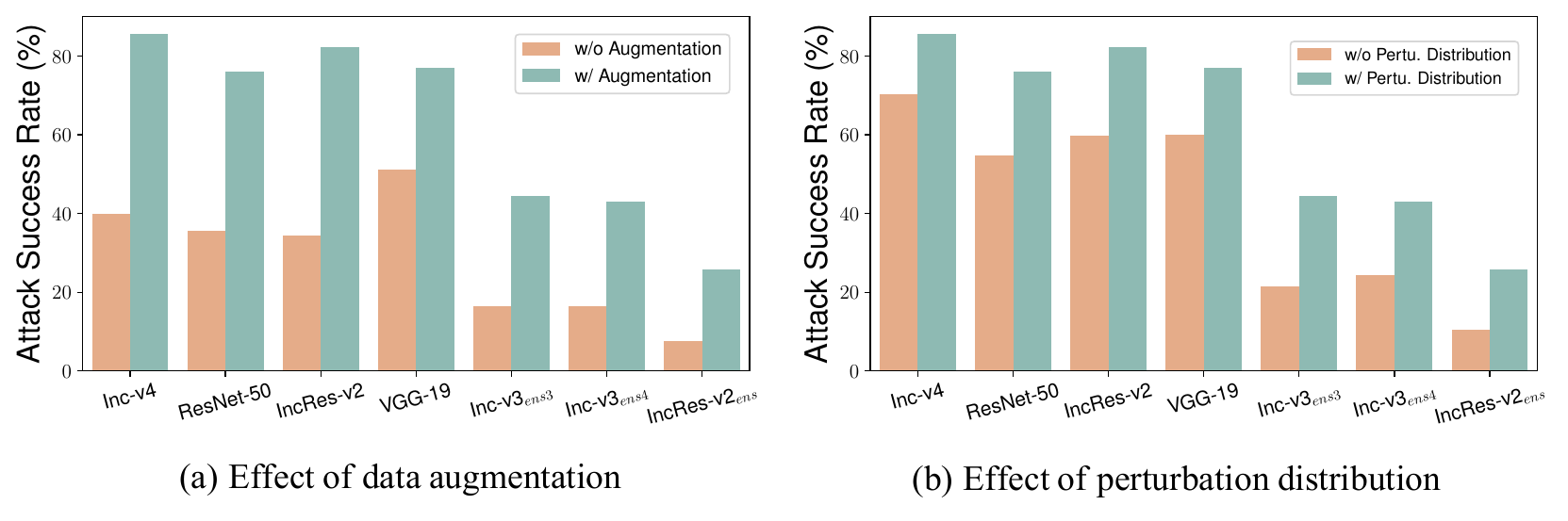}
    \caption{Attack success rates (\%) on seven selected models with adversarial examples generated by ANDA on Inc-v3 model
    }\label{fig:ablation_merge}
\end{figure*}

To determine the contribution of augmentation to ANDA, we compared results with and without augmentation, i.e., $\mathcal{S}$ includes one sample $\{x_{adv}^{(t)}\}$ or $n$ samples $\{\text{AUG}_i(x_{adv}^{(t)})\}_{i=1}^n$, respectively. Figure~\ref{fig:ablation_merge}~(a) illustrates that batch augmentation significantly boosts performance in various black-box models. The main reason is that augmentation introduces stochasticity into the iterative optimization procedure, forming a SGA procedure which help to characterize the asymptotic Gaussian distributions that effectively enhances the robustness of the results.

Furthermore, we analyzed the impact of varying the augmented batch number $n$. Considering both the attack performance and computational overhead, we adjusted $n$ to represent the perfect squares of the number of translated pixels in one dimension, ranging from 4 to 49 (refer detailed augmentation implementation in Section~\ref{sec:augmax}). We observe from Figure~\ref{fig:enable_accu_all} that the overall attack performance generally improves with an increase in $n$. 

\subsection{Effect of Perturbation Distribution}
\begin{figure}[ht]
    \centering
    \includegraphics[width=\linewidth]{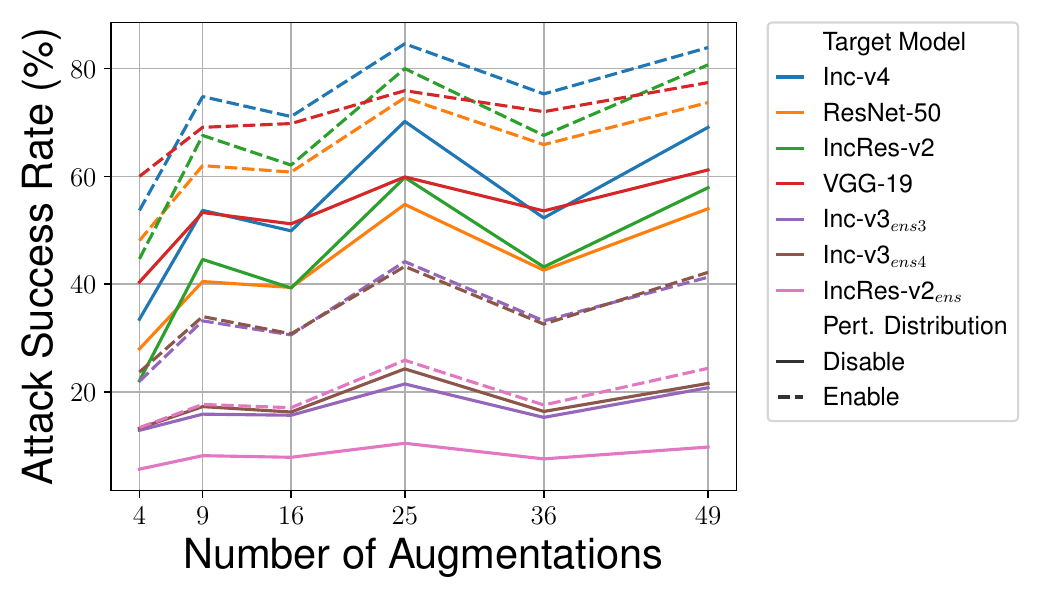}
    \caption{Attack success rates (\%) on seven selected models with adversarial examples generated by ANDA on Inc-v3 model with varying the augmented batch number $n$. Two groups of ASRs, with or without accumulating previous gradients, are shown.
    }
    \label{fig:enable_accu_all}
\end{figure}

We further examined the impact of the approximated perturbation distribution, specifically the accumulation of historical gradients along the optimization trajectory. The results, depicted in both Figure~\ref{fig:ablation_merge} (b) and Figure~\ref{fig:enable_accu_all}, demonstrate that the attack effectiveness is significantly enhanced by the statistical analysis of historical gradients. Notably, the dashed lines in Figure~\ref{fig:enable_accu_all} are consistently higher than their corresponding solid lines of the same color, underscoring the considerable value of incorporating historical gradient data into the optimization process.

\begin{figure}
    \centering
    \includegraphics[width=0.9\linewidth]{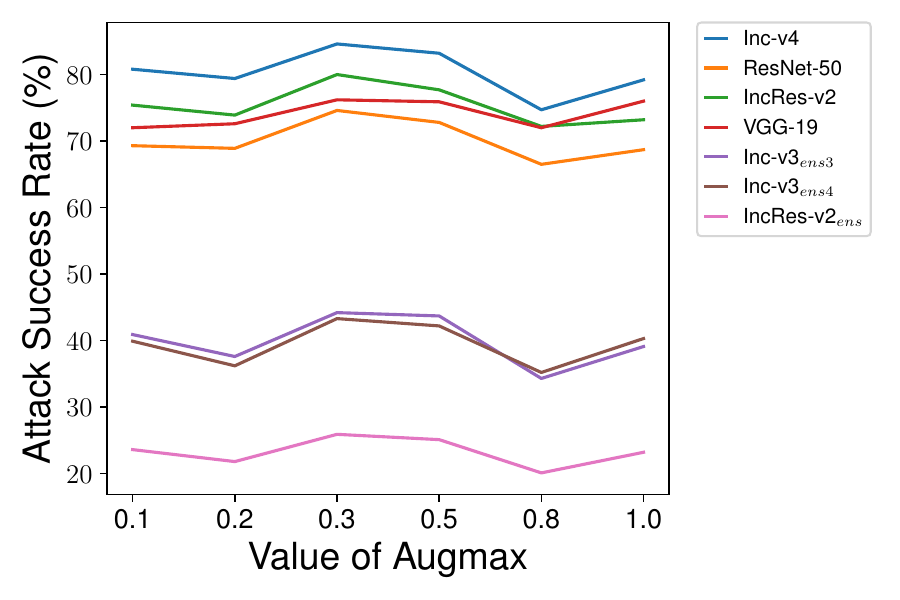}
    \caption{Attack success rates (\%) on the seven models of adversaries generated by ANDA on Inc-v3 model with varying the value of \texttt{Augmax}. 
    }\label{fig:augmax_inc_v3}
\end{figure}

\begin{figure}
    \centering
    \includegraphics[width=\linewidth]{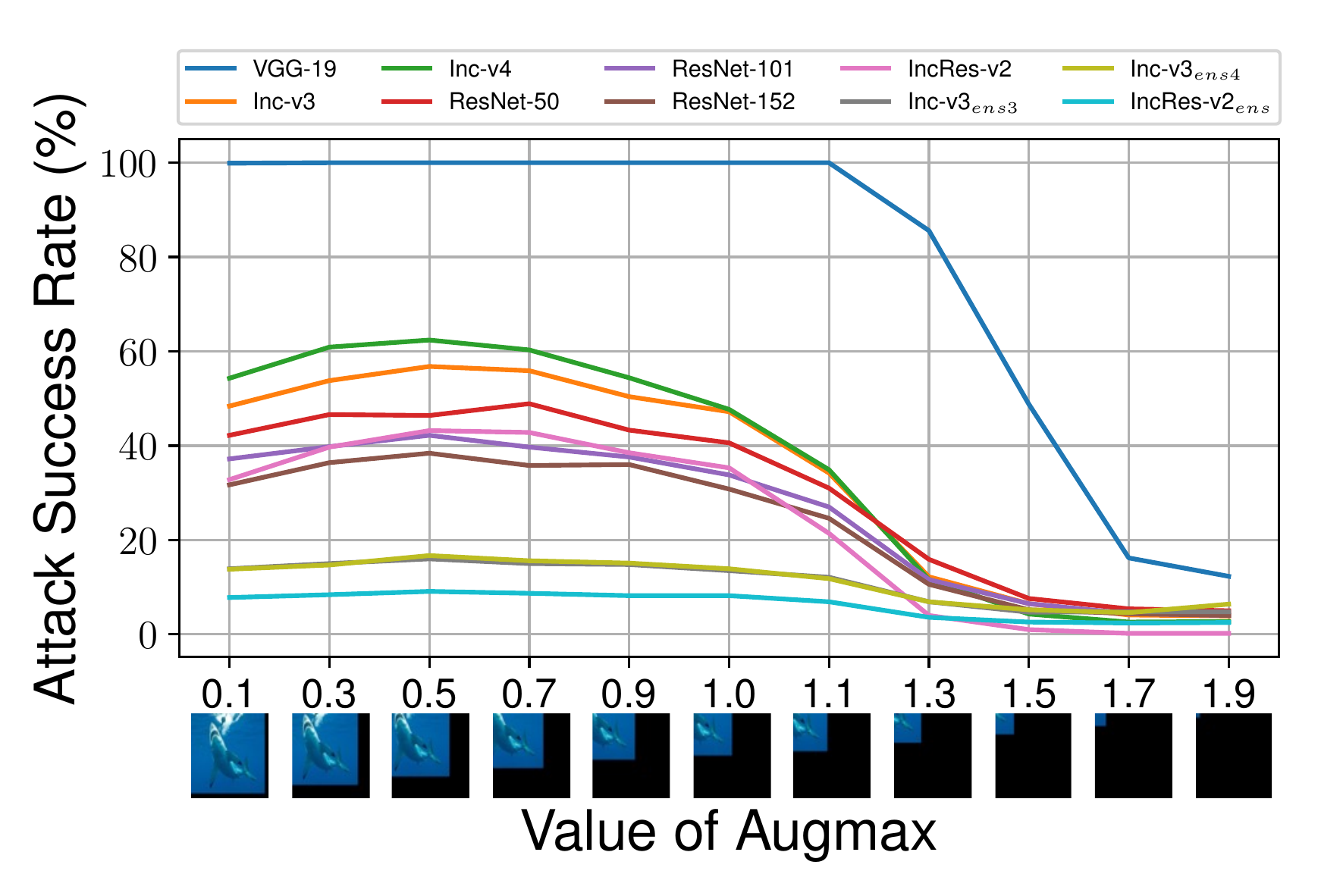}
    \caption{Attack success rates (\%) on target models with adversaries generated by ANDA on VGG-19 model when varying the value of Augmax from $0.1$ to $1.9$.
    }\label{fig:augmax_vgg19_semantic}
\end{figure}

\begin{figure*}[t]
    \centering
    \includegraphics[width=\linewidth]{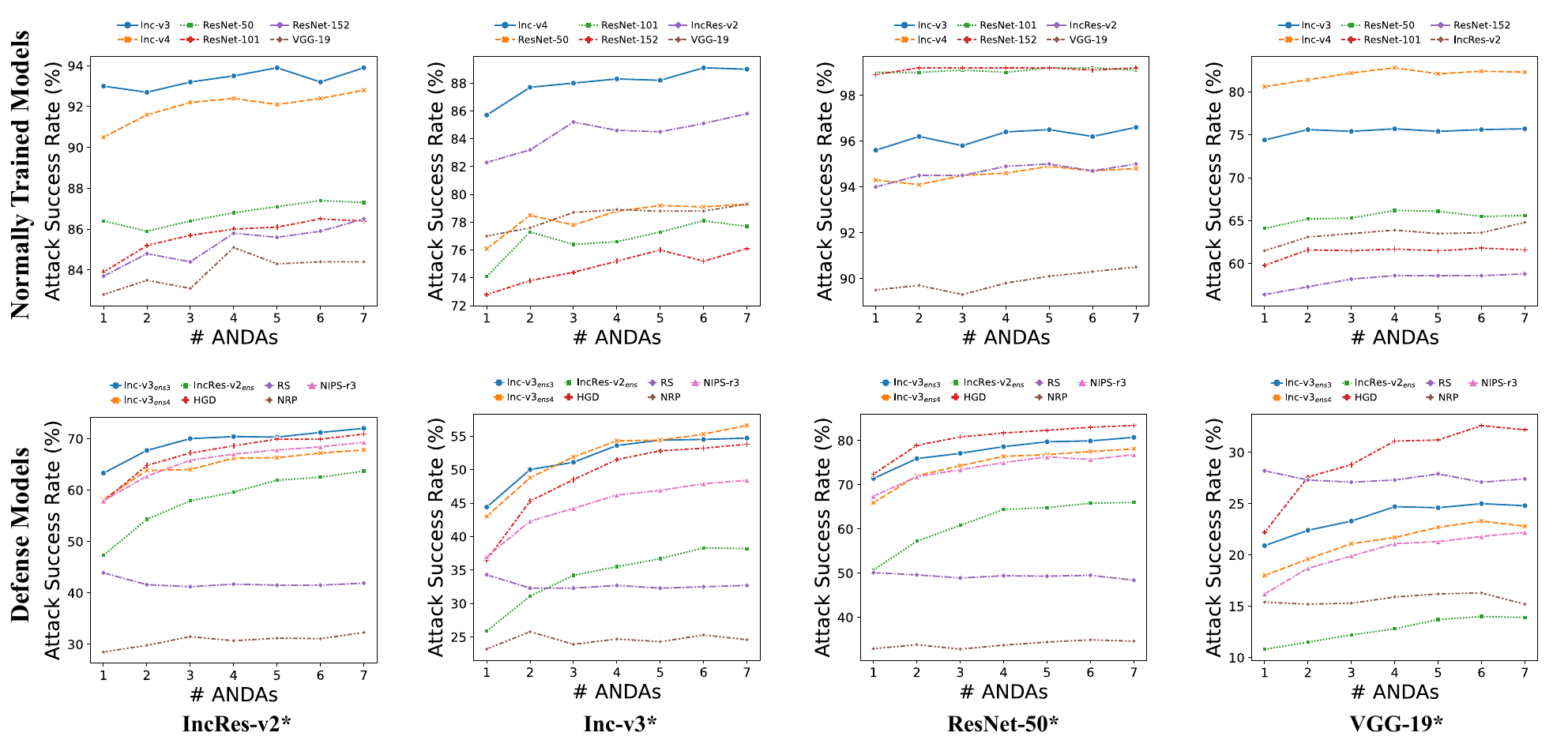}
    \caption{Attack success rates (\%) of normally trained models (top row) and defense models (bottom row) with samples generated by MultiANDA. The results are shown across different source models (indicated by *), with a varying number of ANDAs. Each column corresponds to a specific source model, illustrating the impact of the number of ANDAs on the effectiveness of the attacks.   
    }\label{fig:multianda_hp}
\end{figure*}

\subsection{Magnitude of augmentations}
\label{sec:augmax}

Another important aspect to explore is the extent to which image translations provide the most benefits. For these translations, we utilize $tx$ for the horizontal direction and $ty$ for the vertical direction in the affine transformation, representing the degree of translation. Consider a scenario where we have a source image $x_{src}$ and a translated image $x_{tgt}$. We can then formulate the translation process using the following affine transformation in homogeneous representation. This process involves mapping the $i_{th}$ pixel points from $x_{src}$ to $x_{tgt}$:

\begin{equation*}
    \underbrace{\left[\begin{array}{c}x_i' \\ y_i' \\ 1\end{array}\right]}_{x_{tgt}}=\left[\begin{array}{ccc}1 & 0 & t x \\ 0 & 1 & t y \\ 0 & 0 & 1\end{array}\right]\underbrace{\left[\begin{array}{c}x_i \\ y_i \\ 1\end{array}\right]}_{x_{src}}
\end{equation*}

The magnitudes of $tx$ and $ty$ represent the extent of translations. Specifically, we define two parameters for this process: $n$, representing the number of augmentations, and \texttt{Augmax}, defined as $\max(\text{abs}(tx), \text{abs}(ty))$. It is important to note that the valid range for \texttt{Augmax} is $[0, 2]$, and $n$ takes on values that are perfect squares in our experiments. For example, with $n=25$ and \texttt{Augmax} = 0.3, we evenly distribute $\sqrt{n}$ samples across the interval $(-\texttt{Augmax}, \texttt{Augmax})$ for both $tx$ and $ty$. This results in a set of $tx$ values, denoted as $T_x$, and a set of $ty$ values, denoted as $T_y$. To illustrate, consider $T_x$ as an example:
\begin{equation*}
    \begin{aligned}
        tx_i = -\texttt{Augmax} + i \times \frac{2\texttt{Augmax}}{\sqrt{n} - 1} \\
        T_x = \{tx_i |  i = 0, 1, \ldots, \sqrt{n}-1 \}
    \end{aligned}
\end{equation*}

The process to form $T_y$ follows the same approach as $T_x$. We then calculate the Cartesian product of these two sets to obtain a series of $(tx, ty)$ pairs, which are used for further translation operations. The function $\texttt{AUG}(x)$ represents a set of translated images generated from these operations:

\begin{equation*}
    \begin{aligned}
        T_x \times T_y                        & = \{(tx, ty)| tx \in T_x \wedge ty \in T_y \}                  \\
        \texttt{AUG(x)} = \{ \texttt{AUG}_{i} & (x), \text{where } i \text{ is the index of } T_x \times T_y\}
    \end{aligned}
\end{equation*}

As illustrated in Figure~\ref{fig:augmax_inc_v3}, we present the Attack Success Rates (ASRs) on various models while varying the value of \texttt{Augmax}. The results indicate that image translations within a suitable range can enhance black-box transferability. However, overly intense transformations lead to a loss of intrinsic image information, resulting in decreased attack performance. To illustrate this effect, we extended the range of \texttt{Augmax} and plotted the attack success rates, along with specifically translated images at different values of \texttt{Augmax}, as shown in Figure~\ref{fig:augmax_vgg19_semantic}. It is observed that moderate translation extents, such as $\texttt{Augmax} = 0.3$, are beneficial. Yet, when the value of \texttt{Augmax} exceeds 0.5, there is a significant drop in attack performance, even on a white-box model like VGG-19. This suggests that while appropriate image transformations can promote adversarial transferability, excessive alterations can result in the loss of crucial information.

\subsection{Augmentation Types}

We also empirically explored the impact of different types of augmentations. Translation operations yielded effects similar to random resize and padding, as discussed in ~\cite{Xie19,Zou20}. Although our methods could potentially benefit from stochastic augmentations, this paper focuses solely on deterministic operations, such as translations, to enable more stable quantitative analysis. While scale operations, as in \cite{Lin20}, are feasible, their effectiveness is inherently limited by the nature of the transformation, showing no further improvement beyond $n > 5$ augmentations. Additionally, augmentations like uniform additive noise \cite{Wang21a,Madry18}, Gaussian additive noise\cite{Wu2018}, and Bernoulli multiplicative noise\cite{Wang21b} resulted in only modest performance enhancements in our experiments.

\subsection{Number of Ensemble Models for MultiANDA} 

We investigated the influence of the number of components ($K$) in MultiANDA on performance across our selection of models, including six black-box normally trained models and seven defense models. The results, as shown in Figure~\ref{fig:multianda_hp}, indicate that the performance of MultiANDA steadily improves with an increasing number of ANDAs. This is particularly evident with defense models, where MultiANDA achieves approximately a 10\% improvement in success rates for five of the defense models. These findings further suggest that sampling from a Gaussian mixture distribution enhances sample diversity, thereby boosting the transferability of the perturbed samples.

\end{document}